\documentclass[letterpaper,twocolumn,10pt]{article}
\PassOptionsToPackage{babel=true,verbose=silent,
  spacing=nonfrench,activate={true,nocompatibility},
  final,kerning=true,tracking=true,
  patch=footnote}{microtype}
\usepackage{usenix}

\microtypecontext{spacing=nonfrench}
\usepackage[utf8]{inputenc}
\usepackage[T1]{fontenc}
\usepackage{amsmath}

\usepackage{graphicx}
\usepackage{hyperref}    
\usepackage{cleveref}    

\usepackage{xcolor}
\usepackage{algorithm}
\usepackage{algpseudocode}
\usepackage{microtype}
\usepackage{enumitem}
\usepackage[font=small,labelfont=bf]{caption}
\usepackage{float}
\usepackage{amsmath}
\usepackage{xurl}
\usepackage{xcolor}
\usepackage{xspace}
\usepackage{pifont}
\usepackage[english]{babel}
\usepackage{blindtext}
\usepackage{lipsum}
\usepackage{booktabs}
\usepackage{hhline}
\usepackage{multirow}
\usepackage{soul}

\usepackage{array}
\usepackage{comment}
\usepackage{listings}
\usepackage{dsfont}
\usepackage{xurl}
\usepackage{setspace}
\usepackage[flushleft]{threeparttable}
\usepackage{graphicx}
\usepackage{subcaption}
\usepackage{float}
\usepackage{cuted}     
\usepackage{caption}   
\usepackage{ltablex}
\usepackage{ifluatex}
\usepackage{outlines}
\usepackage{hyperref}
\usepackage{hyperxmp}

\usepackage{algorithm}
\usepackage{algorithmicx}
\usepackage{algpseudocode}
\setlist[enumerate,1]{leftmargin=1.8em} 
\newcommand{\sysname}{RedKnot}
\newcommand{\method}{Elastic Sparsity}
\newcommand{\SegPagedAttention}{SegPagedAttention}

\newcommand{\xhslogofile}{sections/figures/xhs_logo.pdf}
\newcommand{\xhslogo}{%
  \IfFileExists{\xhslogofile}{%
    \includegraphics[height=2.6em]{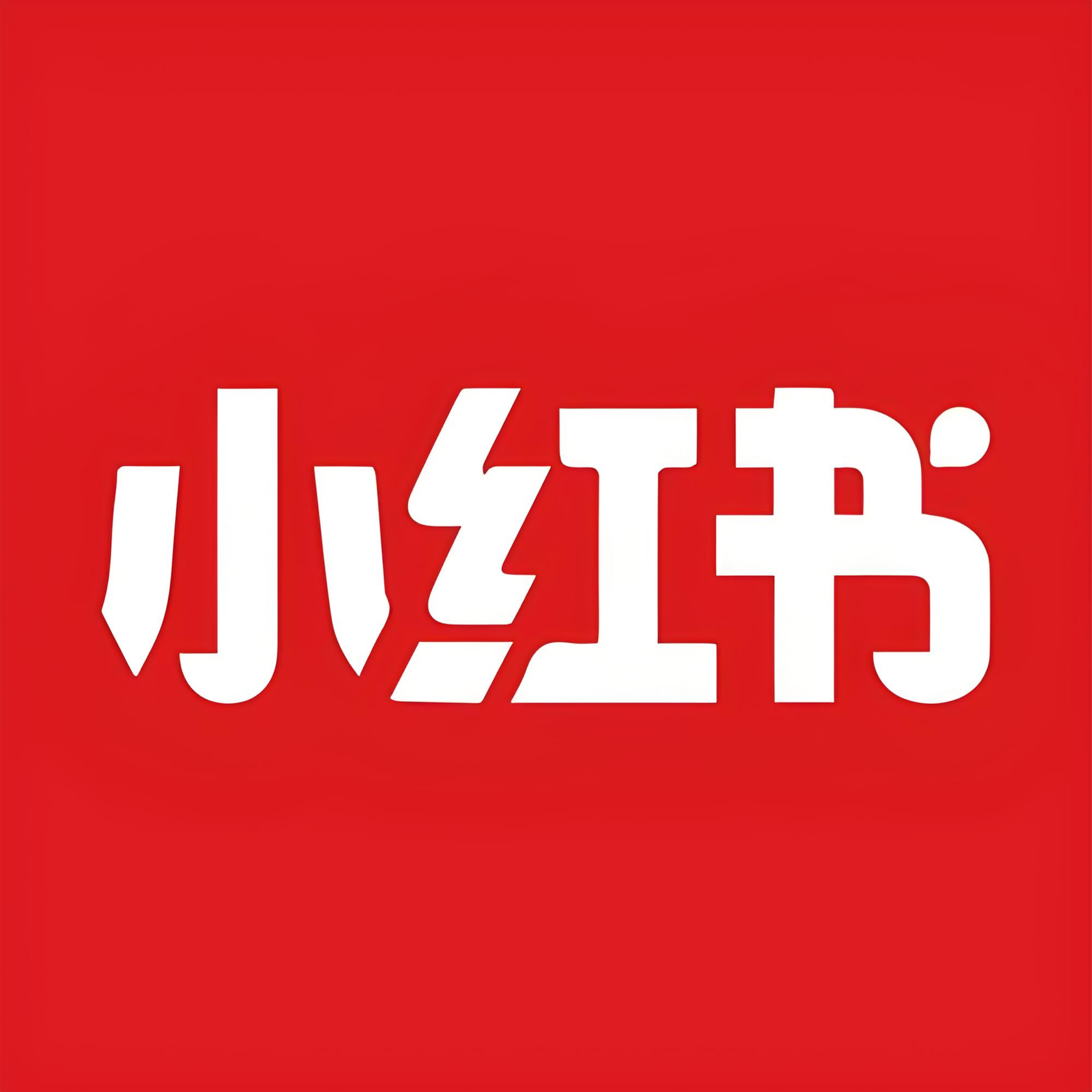}%
  }{%
    \fbox{\rule{0pt}{2.2em}\hspace{0.2em}\scriptsize\itshape logo\hspace{0.2em}}%
  }%
}

\newcommand{\bannertitle}[1]{%
  \begingroup
  \centering
  \noindent\makebox[\linewidth][r]{\smash{\raisebox{0.3em}{\xhslogo}}}\par
  \vspace{-0.6em}%
  {\color{black!75}\rule{\linewidth}{1.1pt}}\par
  \vspace{0.7ex}%
  {\LARGE\bfseries #1\par}%
  \vspace{0.8ex}%
  {\color{black!75}\rule{\linewidth}{1.1pt}}\par
  \endgroup
}
\begin{document}
\newcommand{\edit}[1]{{\color{black} #1}}
\newcommand{\shan}[1]{{\color{red}(Shan: #1)}}
\newcommand{\shanedit}[1]{{\color{red} #1}} 
\newcommand{\hank}[1]{{\color{blue}(Hank: #1)}}
\newcommand{\mm}[1]{{\color{violet}(Michael:#1)}}
\newcommand{\jc}[1]{{\footnotesize\color{orange}{(JC: #1)}}}
\newcommand{\jcedit}[1]{{\color{orange} #1}} 
\newcommand{\yh}[1]{{\footnotesize\color{deepgreen}{(Yuhan: #1)}}}
\newcommand{\jiayi}[1]{{\color{brown}{(Jiayi: #1)}}}
\newcommand{\todo}[1]{{\color{red}{(TODO: #1)}}}
\newcommand{\hcedit}[1]{{\color{black} #1}}
\definecolor{darkkhaki}{rgb}{0.74, 0.72, 0.42}
\newcommand{\hc}[1]{{\color{darkkhaki}{(LHC: #1)}}}
\definecolor{hhcolor}{RGB}{240, 35, 240}
\newcommand{\hh}[1]{{\color{hhcolor}{(Yihua: #1)}}}
\newcommand{\sr}[1]{{\color{cyan!70!blue}{(Siddhant: #1)}}}
\newcommand{\qz}[1]{{\color{purple}{(Qizheng: #1)}}}

\newcommand{\KV}{\ensuremath{KV}\xspace}
\newcommand{\Attention}{\ensuremath{A}\xspace}
\newcommand{\ForwardAttention}{\ensuremath{FA}\xspace}
\newcommand{\LayerIndex}{\ensuremath{i}\xspace}
\newcommand{\TokenIndex}{\ensuremath{j}\xspace}
\newcommand{\UserTokenIndex}{\ensuremath{j'}\xspace}
\newcommand{\ChunkIndex}{\ensuremath{n}\xspace}
\newcommand{\ChunkNum}{\ensuremath{N}\xspace}
\newcommand{\Func}{\ensuremath{F}\xspace}
\newcommand{\Pre}{\textrm{pre}\xspace}
\newcommand{\New}{\textrm{new}\xspace}
\newcommand{\Full}{\textrm{full}\xspace}
\newcommand{\CA}{\ensuremath{C}\xspace}

\newcommand{\KVD}{\ensuremath{\Delta_{\textrm{kv}}}\xspace}
\newcommand{\CAD}{\ensuremath{\Delta_{\textrm{attn}}}\xspace}

\newcommand{\ACA}{{ACA}\xspace}
\newcommand{\HCA}{{HKVD}\xspace}
\newcommand{\name}{\textsc{CacheBlend}\xspace}
\newcommand{\promptcache}{PromptCache\xspace}
\newcommand{\HCF}{HCF tokens\xspace}

\newcommand{\vspacesize}{0.2cm}

\newcommand{\fillme}{{\bf XXX}\xspace}

\newcommand*\circled[1]{\tikz[baseline=(char.base)]{
            \node[shape=circle,fill,inner sep=2pt] (char) {\textcolor{white}{\footnotesize{#1}}};}}

\newcounter{packednmbr}
\newenvironment{packedenumerate}{\begin{list}{\thepackednmbr.}{\usecounter{packednmbr}\setlength{\itemsep}{0.5pt}\addtolength{\labelwidth}{-4pt}\setlength{\leftmargin}{2ex}\setlength{\listparindent}{\parindent}\setlength{\parsep}{1pt}\setlength{\topsep}{0pt}}}{\end{list}}
\newenvironment{packeditemize}{\begin{list}{$\bullet$}{\setlength{\itemsep}{0.5pt}\addtolength{\labelwidth}{-4pt}\setlength{\leftmargin}{2ex}\setlength{\listparindent}{\parindent}\setlength{\parsep}{1pt}\setlength{\topsep}{2pt}}}{\end{list}}
\newenvironment{packedpackeditemize}{\begin{list}{$\bullet$}{\setlength{\itemsep}{0.5pt}\addtolength{\labelwidth}{-4pt}\setlength{\leftmargin}{\labelwidth}\setlength{\listparindent}{\parindent}\setlength{\parsep}{1pt}\setlength{\topsep}{0pt}}}{\end{list}}
\newenvironment{packedtrivlist}{\begin{list}{\setlength{\itemsep}{0.2pt}\addtolength{\labelwidth}{-4pt}\setlength{\leftmargin}{\labelwidth}\setlength{\listparindent}{\parindent}\setlength{\parsep}{1pt}\setlength{\topsep}{0pt}}}{\end{list}}
\let\enumerate\packedenumerate
\let\endenumerate\endpackedenumerate
\let\itemize\packeditemize
\let\enditemize\endpackeditemize

\newcommand{\tightcaption}[1]{\vspace{-0.2cm}\caption{{\normalfont{\textit{{#1}}}}}\vspace{-0.2cm}}
\newcommand{\tightsection}[1]{\vspace{-0.3cm}\section{#1}\vspace{-0.2cm}}
\newcommand{\tightsectionstar}[1]{\vspace{-0.17cm}\section*{#1}\vspace{-0.08cm}}
\newcommand{\tightsubsection}[1]{\vspace{-0.25cm}\subsection{#1}\vspace{-0.1cm}}
\newcommand{\tightsubsubsection}[1]{\vspace{-0.01in}\subsubsection{#1}\vspace{-0.01cm}}

\newcommand{\eg}{{\it e.g.,}\xspace}
\newcommand{\ie}{{\it i.e.,}\xspace}
\newcommand{\etal}{{\it et.~al}\xspace}
\newcommand{\bigO}{\mathrm{O}}
\newcommand{\twlog}{w.l.o.g.\xspac}

\newcommand{\myparashort}[1]{\vspace{0.05cm}\noindent{\bf {#1}}~}
\newcommand{\mypara}[1]{\vspace{0.05cm}\noindent{\bf {#1}:}~}
\newcommand{\myparatight}[1]{\vspace{0.02cm}\noindent{\bf {#1}:}~}
\newcommand{\myparaq}[1]{\smallskip\noindent{\bf {#1}?}~}
\newcommand{\myparaittight}[1]{\smallskip\noindent{\emph {#1}:}~}
\newcommand{\question}[1]{\smallskip\noindent{\emph{Q:~#1}}\smallskip}
\newcommand{\myparaqtight}[1]{\smallskip\noindent{\bf {#1}}~}

\newcommand{\cmark}{\ding{51}}%
\newcommand{\xmark}{\ding{55}}%



\definecolor{backcolour}{rgb}{0.96,0.96,0.96}
\definecolor{codegray}{rgb}{0.5,0.5,0.5}
\definecolor{deepblue}{rgb}{0,0,0.6}
\definecolor{deepred}{rgb}{0.6,0,0}
\definecolor{deepgreen}{rgb}{0,0.5,0}
\lstdefinestyle{mystyle}{
    backgroundcolor=\color{backcolour},   
    commentstyle=\color{codegreen},
    morekeywords={self, True},
    keywordstyle=\color{deepblue},
    numberstyle=\tiny\color{codegray},
    emph={MyClass,__init__,EncodingType,Image},
    emphstyle=\color{deepred},
    stringstyle=\color{deepgreen},
    basicstyle=\ttfamily\footnotesize,
    breakatwhitespace=false,         
    breaklines=true,                 
    captionpos=b,                    
    keepspaces=true,                 
    numbers=left,                    
    numbersep=5pt,                  
    showspaces=false,                
    showstringspaces=false,
    showtabs=false,                  
    tabsize=1
}

\title{\bannertitle{RedKnot: Efficient Long-Context LLM Serving with\\ Head-Aware KV Reuse and SegPagedAttention}}
 \author{
 {\rm Yang Liu}\textsuperscript{$\ast$}$^\dagger$ \quad
 {\rm Zhaokai Luo}\textsuperscript{$\ast$}$^\dagger$\quad
 {\rm Huayi Jin}$^\dagger$ \quad
 {\rm Ruozhou He}$^\dagger$ \quad
 {\rm Chenchen Hong}$^\dagger$ \quad
 {\rm Zhiyong Wang}$^\dagger$ \quad\\
 {\rm Boyu Wang}$^\mathparagraph$ \quad
 {\rm Guanjie Chen}$^\mathparagraph$ \quad
 {\rm Yifei Liu}$^\dagger$ \quad
 {\rm Tao Xie}$^{\mathsection\ddagger}$ \quad
 {\rm Junhao Hu}$^{\mathsection\ddagger}$\\
 $^\dagger${\it Xiaohongshu Inc., China} \quad
$^\ddagger${\it Peking University}\quad
$^\mathparagraph${\it Huawei Cloud}
\\
$^\mathsection${\it Beijing Tongming Lake Information Technology Application Innovation Center}\\[0.6ex]
 {\normalsize\rm
   \textsuperscript{$\ast$}Corresponding to:\;
   Yang Liu~\href{mailto:xiaoyi52@xiaohongshu.com}{\textless xiaoyi52@xiaohongshu.com\textgreater} or\;
   ZhaoKai Luo~\href{mailto:luozhaokai@xiaohongshu.com}{\textless luozhaokai@xiaohongshu.com\textgreater}%
 }\\
}
\maketitle
\vspace*{1ex}
\thispagestyle{plain}
\pagestyle{plain}
\begin{abstract}
As the input length of large language model (LLM) serving continues to grow, the KV cache has become a dominant bottleneck in AI infrastructure. ~It limits GPU memory capacity, serving concurrency, cache reuse, and distributed scalability. ~Multiple important problems, including position-independent KV cache, prefix KV cache compression, hot/cold KV cache separation, and distributed KV cache management, all depend on how the KV cache is represented and managed. ~\textbf{However}, existing serving systems largely rely on a monolithic KV cache abstraction, where the KV cache is treated as a homogeneous sequence of token-level memory blocks and managed with similar policies across attention heads and serving scenarios. We observe that KV cache utility is highly structured across KV heads: different heads exhibit different functional roles, attention distances, and runtime importance. Therefore, a full KV cache is not always necessary for every head, token range, or serving scenario.\\
\indent We present \sysname, a head-aware KV cache management system for LLM serving. ~\textbf{\sysname\ breaks the conventional monolithic KV cache abstraction by decomposing the KV cache along KV heads, whose importance and effective attention ranges vary significantly across serving scenarios}. ~This head-level decomposition turns the KV cache from a monolithic tensor abstraction into a structured memory object, enabling \sysname\ to uniformly support position-independent KV reuse, prefix KV compression, hot/cold KV separation, and distributed KV placement while preserving output fidelity and improving resource efficiency, without requiring model retraining or fine-tuning. \sysname\ establishes a new foundation for AI infrastructure by transforming the KV cache from a monolithic, passive runtime artifact into a dynamic, model-aware runtime substrate for scalable LLM serving.
\end{abstract}

\def\thefootnote{\dag} 
\pagestyle{empty}
\pagestyle{plain}
\section{Introduction}
\label{sec:intro}
\begin{figure}[!t]
  \centering
  \vspace{8ex}
\includegraphics[width=0.99\linewidth]
{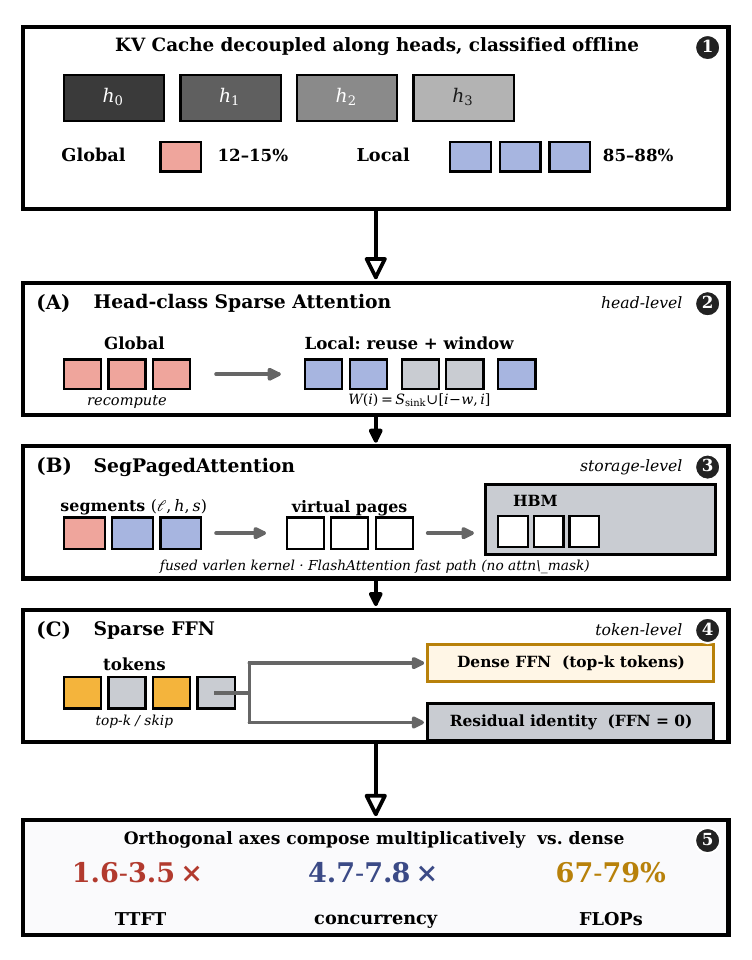}
  \caption{\sysname~decouples the KV cache along
the head dimension, classifies heads into global and local classes, and
co-optimizes sparse attention,~sparse FFN execution with selected tokens and SegPagedAttention. ~The combined
design yields 1.6--3.5$\times$ lower TTFT, 4.7--7.8$\times$ higher
concurrency, and 67--79\% fewer FLOPs compared with dense attention.}
\vspace{-3ex}
  \label{fig:introduce}
\end{figure}
Large language models (LLMs) have become the execution substrate of
modern software systems. ~Retrieval-augmented generation (RAG)~\cite{lewis2020rag,gao2024rag_survey} routinely concatenates
tens of thousands of retrieved tokens into a single prompt; coding
agents such as Claude Code~\cite{anthropic2025claudecode}, OpenAI Codex~\cite{openai2025codex}, and
OpenClaw~\cite{openclaw2025} chain dozens of tool calls whose aggregated inputs reach
hundreds of thousands of tokens; and long-horizon agent frameworks fold memory, tool outputs, and retrieval results into a single context. ~In all three regimes, input length grows far faster than GPU memory bandwidth and arithmetic throughput improve, making the time-to-first-token (TTFT) of the prefill phase the dominant cost of interactive serving---a single $64$\,K-token prompt on a $4\times$ tensor-parallel Llama-3.3-70B deployment takes roughly $64$\,s under vanilla dense
attention~\cite{kwon2023vllm,dao2023flashattention2,dao2024flashattention3}, a latency incompatible with any interactive SLO.~Two parallel research threads have emerged in response. ~First, position-independent KV
caching~\cite{yao2024cacheblend,hu2025epic,wang2026prophetkv,gim2024promptcache} (PIC), amortizes prefill across requests that share document chunks
by precomputing reusable key-value (KV) entries and splicing them into subsequent prompts regardless of positional shifts. Second, multi-head KV
sparsification~\cite{xiao2025duoattention,tang2025razorattention,fu2025headkv,zhang2023h2o,liu2024scissorhands,lin2025compresskv}, exploits the observation that only a small fraction of attention
heads requires full-sequence access, evicting or compressing the KV of the remaining heads. 

Both directions promise large savings on paper, yet deployed systems realize only a fraction of the predicted speedup. ~We trace this gap to three structural mismatches between current
implementations and the sparsity actually present in the workload.
~First, existing PIC schemes recover the residual error of cached KV at the \emph{token} granularity, but the underlying sparsity is \emph{per-head}: different heads attend to different subsets of tokens, so any token-level selector that satisfies every head must take their union, which routinely covers a large fraction of the chunk and defeats
reuse~(\cref{sec3.1}). ~Second, PIC implicitly assumes attention dominates prefill, yet at the $2$--$8$\,K segment lengths characteristic of agent workloads the feed-forward network (FFN)
contributes $57$--$62\%$ of TTFT
(\cref{fig:ffn_share_combined})---a cost no attention-side technique can touch. ~Third, even when an algorithm retains only a fraction of KV
per head, the KV is still stored in the dense \texttt{[B,H,L,D]} layout as in PagedAttention~\cite{kwon2023vllm,zheng2024sglang} and expressed at runtime via an \texttt{attn\_mask}, which disables the FlashAttention fast path under standard SDPA dispatch and incurs a $4.9$--$7.6\times$ kernel penalty~(\cref{sec3.4}); algorithmic byte savings therefore never materialize as compute savings. ~The three failures share a
single root: \emph{the recovery, compute, and storage granularities of existing PIC systems do not match the per-head and per-channel sparsity structure of the workload}, and closing the gap requires
aligning all three axes simultaneously.

This paper presents \textbf{\sysname}, a serving system that operationalizes this alignment through three co-designed mechanisms, illustrated in \cref{fig:introduce}. \ding{182}~\emph{Head-class
sparsification} classifies every $(layer, head)$ pair offline as
either ~\ding{183}~global ($12$--$15\%$ of heads, re-prefilled on reuse) or local ($85$--$88\%$, reused verbatim within a sliding window), with
an adaptive runtime restore that promotes individual local heads to full attention when an edge-mass signal detects insufficient classification---eliminating the cascading staleness of token-level
patches. ~\ding{184}~\emph{SegPagedAttention} replaces the dense layout with a per-$(layer, head)$ paged KV store and a fused varlen attention
kernel that physically retains only the tokens each head needs, keeping every head on the FlashAttention fast path without ever constructing an \texttt{attn\_mask} and yielding kernel speedups
that \emph{rise monotonically} with context length, the opposite of the diminishing returns of dense+mask implementations. ~\ding{185}~\emph{Sparse
FFN} evaluates only the top-$k$ tokens with the highest attention scores, an axis structurally independent of context length and therefore the only lever that accelerates the short-context agent workloads that attention-side optimization leaves untouched. ~Because the three mechanisms operate on orthogonal axes (heads, storage, channels), their savings compose multiplicatively rather than competing for the same slack.\\
\indent We implement \sysname\ on top of SGLang~\cite{zheng2024sglang}, and evaluate it on an $8\times$ NVIDIA H800 (80\,GB) server across three models (Mistral-7B, Qwen3-32B, Llama-3.3-70B), six QA datasets, and context lengths from $8$\,K to $128$\,K. ~\ding{186}~Across this sweep, \sysname\ ~delivers up to $1.6-3.54\times$ TTFT speedup and $4.7$--$7.8\times$ more concurrent sessions per GPU while cutting prefill FLOPs by $67-79.5\%$, with end-to-end accuracy matching or exceeding the dense baseline. ~We will continue to maintain our engineering project in the open-source community at~\url{https://github.com/rednote-machine-learning/RedKnot}.~The experimental results in the paper are for reference only, the test results from the open-source community code shall prevail. ~Our contributions are summarized as follows:
\begin{itemize}
  \item We propose \textbf{head-class sparsification}, a high-fidelity low-cost PIC scheme that recovers cached KV at the granularity of attention heads rather than tokens, paired with selected \textbf{token-level Sparse FFN} to attack the short-context FFN bottleneck that no attention-side technique can touch.
  \item We design and implement \textbf{SegPagedAttention}, a per-$(layer, head)$ paged KV store backed by a fused varlen attention kernel that physically materializes per-head sparsity and keeps every head on the FlashAttention fast path, eliminating the $4.9$--$7.6\times$ \texttt{attn\_mask} penalty of dense+mask implementations and turning algorithmic byte savings into kernel-level speedups that scale with context length.
  \item We implement \sysname ~as an \textbf{architecture-agnostic serving} runtime built around architecture-dependent reusable states.
\end{itemize}

\section{Background}\label{background} 
In this section, we review two lines of work that motivate our design. 
First, position-independent KV-cache reuse enables systems to amortize prefill computation across prompts even when reusable text chunks do not appear after the same prefix. 
Second, structured sparsity in attention and FFNs shows that different model components may require different amounts of context and computation during inference. 
Together, these observations suggest that KV-cache management should be both chunk-aware and structure-aware, rather than treating all tokens, layers, and heads uniformly.

\subsection{Position-Independent KV Cache}

Many emerging workloads, such as retrieval-augmented generation (RAG), long-context question answering, few-shot learning, and agentic applications, repeatedly use the same text chunks, but these chunks do not always appear after the same prefix. This motivates position-independent KV cache (PIC), where the system attempts to reuse precomputed KV cache even when the reused text appears at different positions in the final prompt.

~CacheBlend~\cite{yao2024cacheblend} is a representative work in this direction. It observes that reusable text chunks in RAG workloads may appear after different prefixes, where their precomputed KV caches cannot be directly reused because they miss cross-attention with the preceding texts. ~It addresses this problem by blending precomputed KV caches and selectively recomputing a subset of tokens to recover output quality. ~EPIC~\cite{hu2025epic} further formalizes position-independent caching and introduces a serving system that enables modular KV cache reuse regardless of token chunk positions. It mitigates the attention-sink effect introduced by independently cached chunks and improves TTFT and throughput with negligible or no accuracy loss.~ProphetKV~\cite{wang2026prophetkv} focuses on long-context RAG and improves selective recomputation by prioritizing tokens according to their semantic relevance to the user query, addressing the crowding-out effect where globally salient but query-irrelevant tokens consume the limited recomputation budget. ~In parallel, CacheSlide~\cite{liu2026cacheslide} studies non-prefix KV cache reuse in agentic workloads, where previously generated contexts can be reused across repeated execution paths. Its reuse model, however, assumes a relatively fixed order among reusable KV states. This assumption differs from typical PIC settings, such as RAG and multi-agent applications, where reusable chunks are often retrieved, reordered, and composed dynamically. Therefore, CacheSlide addresses an important but different point in the design space of non-prefix KV cache reuse.

\subsection{Structured Sparsity in Attention and FFNs}\label{sec:algorithmic_sparsity}

Modern LLM inference contains two major computational components: grouped-query attention (GQA)~\cite{ainslie2023gqa,shazeer2019fast,touvron2023llama,jiang2023mistral,dubey2024llama}
and feed-forward networks (FFNs)~\cite{vaswani2017attention,liu2023dejavu,song2024powerinfer}. ~Recent studies show that both components exhibit strong runtime sparsity, suggesting that dense computation is not always necessary for every head, token, or neuron.\\
\indent \textbf{\emph{Multi-head attention sparsity.}}
~Long-context attention exhibits strong and relatively stable head-level heterogeneity. ~Existing studies suggest that whether an attention head behaves as a global/retrieval head or a local/streaming head is largely a model-intrinsic property associated with its layer and head index, rather than being determined from scratch for each input context. ~In other words, once a specific $(\text{layer}, \text{head})$ is identified, its effective attention scope can often be profiled offline and reused across requests.~StreamingLLM~\cite{xiao2024streamingllm} identifies the attention sink phenomenon, where several initial tokens receive high attention even when they are not semantically important. By preserving these sink tokens and recent tokens, StreamingLLM supports stable streaming generation with a bounded KV cache. DuoAttention~\cite{xiao2025duoattention} further separates attention heads into retrieval heads and streaming heads: retrieval heads require full-context access, while streaming heads mainly attend to recent tokens and attention sinks and can use a constant-length KV cache.~RazorAttention~\cite{tang2025razorattention} similarly observes that most heads primarily focus on local context, while only a small number of retrieval heads need access to the full KV cache.~MInference~\cite{jiang2024minference} shows that long-context attention heads follow distinct sparse patterns, such as A-shape, Vertical-Slash, and Block-Sparse patterns, and assigns different sparse computation strategies to different heads during prefill.\\
\indent These works suggest that attention computation is naturally structured across heads. ~More importantly, this structure can be represented as a stable per-head cache requirement. ~For example, a profiled local head at a given layer may only need a bounded window of recent tokens plus a small number of sink tokens, while a profiled global head may require full-context KV access. ~Therefore, different heads may require different context ranges and KV cache residency policies, and such policies can be determined at the granularity of ~$(\text{layer}, \text{head})$ rather than recomputed from scratch for every request.\\
\indent \textbf{\emph{FFN sparsity.}}
In parallel, FFN computation also exhibits strong activation sparsity. ~DejaVu~\cite{liu2023dejavu} observes contextual sparsity in Transformer inference, where only a small input-dependent subset of attention heads and MLP parameters is needed to approximate the dense model output. ~PowerInfer~\cite{song2024powerinfer} exploits the power-law distribution of neuron activations in LLM inference, separating frequently activated hot neurons from cold neurons to reduce GPU memory demand and CPU-GPU data movement. More recent activation-sparsity methods, such as CATS~\cite{lee2024cats}, TEAL~\cite{liu2025teal}, and ProSparse~\cite{song2025prosparse}, further show that FFN or activation sparsity can be induced or exploited in modern LLMs to reduce computation with limited quality degradation.~Together, these studies show that LLM inference contains structured sparsity beyond token-level reuse. ~For multi-head attention, KV-cache heads can be largely summarized as global heads and local heads: global heads require long-range KV access, while local heads mainly rely on recent tokens, attention sinks, or bounded windows, enabling substantial cache reduction with negligible accuracy loss. ~FFNs further expose activation-level sparsity, where only a subset of neurons or channels dominates computation for a given input. 
\subsection{Heterogeneous KV Cache and Runtime-State Architectures}\label{sec:hybirdattention&MLA}
Modern long-context large language models are moving beyond homogeneous decoder-only Transformer architectures. Early LLM serving systems were mostly designed around multi-head attention (MHA), multi-query attention (MQA), or grouped-query attention (GQA). In these architectures, the runtime history state is typically represented as an explicit key-value (KV) cache, indexed by layer, KV head, and token position.~However, recent model families increasingly integrate softmax attention, recurrent linear attention, latent attention, and sparse Mixture-of-Experts (MoE) feed-forward networks within a single architecture, leading to a stronger trend toward architectural heterogeneity. Accordingly, the runtime history exposed by these models no longer takes a unified form as a traditional explicit KV cache. Instead, it appears in multiple forms, including explicit KV tensors in standard attention layers, compressed latent states in Multi-head Latent Attention (MLA), and fixed-size recurrent states in linear attention layers.~Next, we introduce two representative architectures that expose heterogeneous KV-cache states.\\
\indent \textbf{\emph{Hybrid Attention and MoE Architectures.}}~
Recent long-context LLMs are increasingly adopting hybrid architectures that combine different sequence-mixing mechanisms within the same model.~Instead of using full softmax attention in every layer, these models interleave full-attention layers with recurrent or linear-attention layers, and then attach sparse Mixture-of-Experts (MoE) feed-forward blocks after each sequence-mixing module. ~Full-attention layers preserve global token-to-token interaction and are especially important for retrieval, long-range dependency modeling, and reasoning over distant evidence. Linear-attention layers, in contrast, summarize the past into a bounded recurrent state, which reduces the cost of processing long contexts because the model does not need to explicitly attend to all previous tokens at every layer. ~Sparse MoE layers further reduce the per-token feed-forward cost by activating only a small subset of experts for each token~\cite{shazeer2017outrageously,lepikhin2021gshard,fedus2022switch,dai2024deepseekmoe}

\indent A representative example is the \emph{Qwen3.5 model family}~\cite{qwen35_blog,qwen35_collection}.~Qwen3.5 adopts a hybrid attention--MoE design that mixes Gated DeltaNet linear-attention layers with full-attention layers, and places sparse MoE blocks after the sequence-mixing modules~\cite{qwen35_35b_a3b_modelcard,qwen35_397b_a17b_modelcard,yang2025gateddeltanet}. For example, Qwen3.5-35B-A3B follows this hybrid pattern at a medium scale, while Qwen3.5-397B-A17B scales the same design to a much larger MoE model with 397B total parameters and 17B activated parameters~\cite{qwen35_397b_a17b_modelcard}. ~This design creates a heterogeneous runtime state. The full-attention layers still expose explicit KV cache, similar to conventional Transformer layers. In contrast, the linear-attention layers maintain compact recurrent states that summarize previous tokens. Therefore, the reusable history in such models is no longer a single dense KV-cache tensor. It consists of both explicit KV states from full-attention layers and recurrent states from linear-attention layers. This observation motivates a more general serving abstraction that manages heterogeneous runtime states rather than only token-level KV blocks.\\
\indent \textbf{\emph{Compressed Muti-Head Attention Architectures.}}~
Another important direction in modern long-context LLMs is to replace explicit per-head KV cache with compressed attention states. ~A representative example is the \emph{DeepSeek model family}. ~DeepSeek-V2 introduced \emph{Multi-head Latent Attention (MLA) } to reduce KV-cache memory while preserving the modeling capacity of multi-head attention~\cite{deepseekv2}. ~DeepSeek-V3 further adopted MLA together with DeepSeekMoE, showing that latent attention states and sparse MoE feed-forward networks can be combined to improve inference efficiency and training economy at large scale~\cite{deepseekv3,dai2024deepseekmoe}.~The key idea of MLA is to cache a compressed latent representation of keys and values instead of storing all per-head KV tensors explicitly. During inference, the model reconstructs the head-specific key and value representations from this latent state when computing attention. To preserve positional information, MLA also keeps a decoupled RoPE-related component. As a result, the runtime state of an MLA layer is no longer a dense tensor indexed directly by layer, head, and token position. Instead, it consists of a compressed latent KV stream together with lightweight position-dependent side information. This design substantially reduces the physical cache footprint, but it also changes the cache object that a serving system must manage.\\
\indent DeepSeek-V4 continues this trend toward compressed long-context attention. The DeepSeek-V4 series, including DeepSeek-V4-Pro~\cite{deepseekv4_pro_modelcard} and DeepSeek-V4-Flash~\cite{deepseekv4_flash_modelcard}, targets million-token contexts and introduces a hybrid compressed-attention design that combines Compressed Sparse Attention (CSA) and Heavily Compressed Attention (HCA) for long-context efficiency. ~Although the detailed attention stack differs from the original MLA formulation in DeepSeek-V2/V3, the serving implication is similar: long-range history is not naturally represented as ordinary per-head KV blocks. Instead, the runtime must manage compressed latent states, sparse extra-cache states, and local sliding-window states. Optimized kernels such as FlashMLA further expose this compressed-attention execution model to serving systems~\cite{deepseek_flashmla}.

\section{Motivation and Opportunity}\label{sec3}
In this section, we analyze why existing position-independent KV cache (PIC) systems leave most of their promised speedup unrealized, and we identify the structural opportunities that motivate \sysname's design.
\subsection{Limitations of Token-Level Recovery}\label{sec3.1}
\begin{figure}[t]
  \centering
\includegraphics[width=1\linewidth]{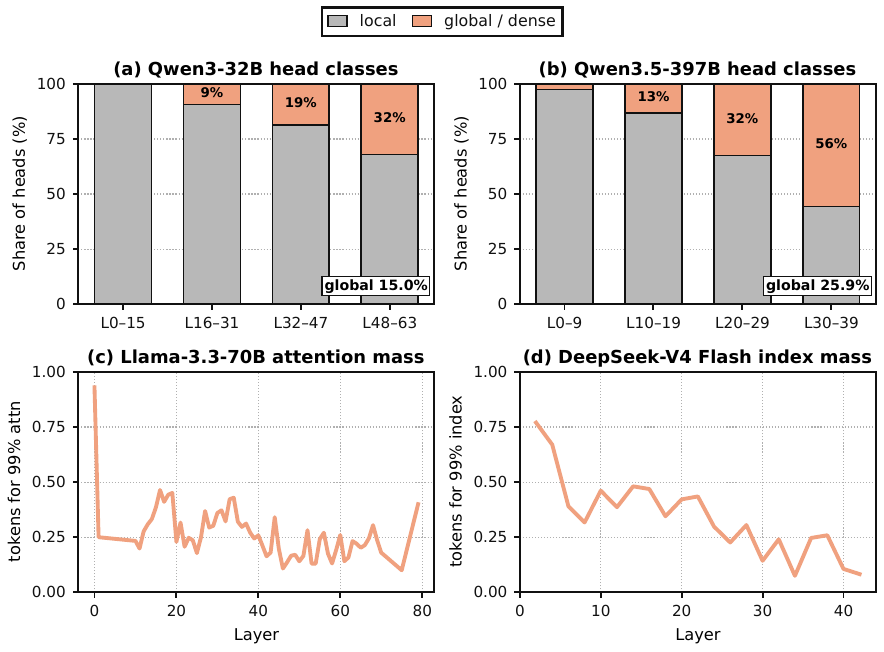}
  \caption{Subfigures (a) and (b) demonstrate that the globality of the KV cache increases markedly with layer depth, revealing a prevalent phenomenon in which locality dominates in shallow layers and globality dominates in deep layers. Subfigures (c) and (d) show that, after the attention computation, the token-level importance becomes non-uniform; as the number of layers grows, attention increasingly concentrates on a small subset of tokens, whereas it remains dispersed in shallow layers.}
  \vspace{-4ex}
  \label{fig:global-local-heads-2}
\end{figure}
\begin{figure}[t]
  \centering
\includegraphics[width=0.99\linewidth]{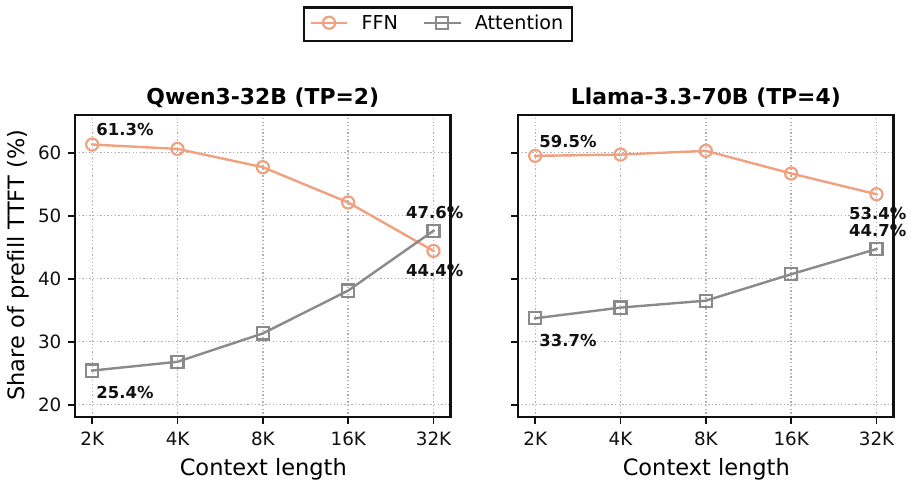}
  \caption{For short contexts, prefill TTFT is dominated by FFN computation rather than KV-cache construction.}
  \vspace{-1ex}
  \label{fig:ffn_share_combined}
\end{figure}

Existing PIC systems, such as CacheBlend, EPIC, and ProphetKV, attempt to exploit sparsity of attention~\cite{xiao2025duoattention,tang2025razorattention,jiang2024minference,fu2024moa} by selecting a subset of tokens for recomputation or correction. ~The underlying assumption is that only a small number of tokens are critical for recovering the quality gap between reused KV cache and full prefill. However, this token-level sparsity becomes less effective under multi-head attention. Different heads may attend to different token subsets and exhibit different position sensitivity. ~As a result, the tokens that need correction for one head may differ from those required by another head. ~When the system must recover all heads of a selected token together, the effective recomputation set becomes the union of head-specific important tokens. \\
\indent~This union can cover a large portion of the chunk, forcing the system to recompute many tokens to recover accuracy. ~This creates a fundamental limitation for token-level PIC recovery. ~\textbf{Even if each individual head is sparse, their important token sets may be diverse across heads. ~Token-level methods cannot exploit this per-head sparsity because they collapse all heads of a token into one recovery decision.} ~As shown in Figure~\ref{fig:global-local-heads-2}, we evaluate the datasets described in Section~\ref{sec:eval:setup} with Qwen3-32B and Qwen3.5-397B. We compute, for each KV-cache head, the set of sparse tokens and then take the union across heads. The reported value is the ratio of the number of tokens in this union to the total number of input tokens. Figures~\ref{fig:global-local-heads-2}~(a) and~\ref{fig:global-local-heads-2}~(b) show that this union covers nearly all input tokens in shallow layers, yielding a ratio close to 1. In deeper layers, the coverage of sparse tokens across heads drops substantially.~Therefore, they face an unfavorable trade-off: selecting fewer tokens may leave some head-specific errors uncorrected and hurt output quality, while selecting more tokens improves fidelity but quickly reduces the TTFT benefit of KV reuse.
Beyond the attention-level bottleneck, existing PIC systems also overlook a complementary challenge that arises in short-context scenarios. ~\emph{For short input sequences, the feed-forward network (FFN) computation dominates the time-to-first-token (TTFT) rather than attention.} ~As shown in Figure~\ref{fig:ffn_share_combined}, FFN layers consistently dominate TTFT across both Qwen3-32B and Llama-3.3-70B, accounting for over 57\% of total prefill time at short contexts (2K--8K tokens). ~As context length increases, the FFN share gradually declines while the attention share rises, reflecting the quadratic scaling of attention with sequence length. ~Notably, even at 32K tokens, FFN still contributes 44.4\% and 53.4\% of prefill
TTFT for Qwen3-32B and Llama-3.3-70B, respectively, underscoring FFN computation as the primary bottleneck for long-context RAG prefill workloads. ~In this regime, even a perfect KV cache reuse strategy that eliminates all attention recomputation yields only marginal TTFT reduction, because the FFN
cost remains unaddressed. ~From a layer-level perspective on token-level sparsity, as shown in Figure~\ref{fig:global-local-heads-2}~(c) and ~(d), we observe that \emph{as the layer depth increases, attention becomes increasingly concentrated on a smaller subset of tokens.}~As a result, conventional PIC implementations may still suffer from substantial TTFT in short-context settings, which limits the extension of PIC from RAG-centric workloads to broader application domains, such as the widely adopted agent-based scenarios. ~\textbf{Reducing the FFN computation cost can substantially improve the generality of PIC systems, enabling their benefits to extend beyond long-context RAG workloads}.
\subsection{Head-Level Sensitivity in Attention}\label{sec3.2}
\begin{figure}[t]
  \centering
\includegraphics[width=0.98\linewidth]{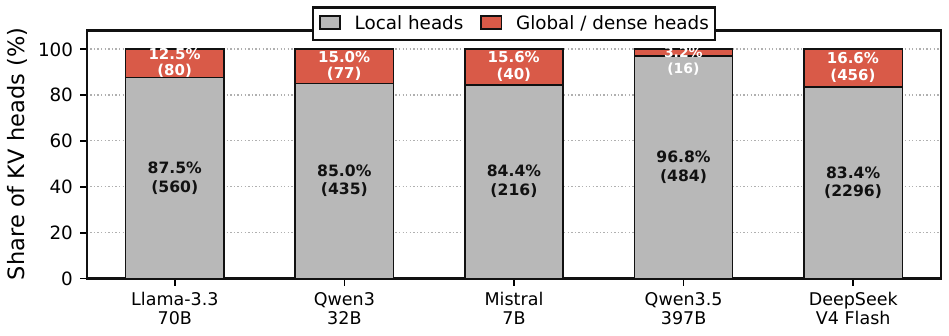}
  \caption{Number of global and local KV cache heads across representative models.}
  \vspace{-3ex}
  \label{fig:global-local-heads}
\end{figure}
Position-independent KV cache (PIC) aims to reuse the precomputed KV cache of a text chunk even when the chunk appears after different prefixes. Compared with full prefill over the newly composed prompt, different KV heads can exhibit substantially different degrees of deviation from their cached states. We observe that some heads change significantly because their attention is strongly affected by the preceding prefix, while others remain nearly unchanged because their attention is mostly confined to local context. Following the discussion in Section~\ref{sec:algorithmic_sparsity}, we define the former as \emph{global heads} and the latter as \emph{local heads}.~This distinction is useful because the global and local behavior of a KV head is largely stable for a given layer and head index. ~Prior studies have shown that only a small fraction of heads require long-range or full-context access, while most heads behave as local or streaming heads~\cite{xiao2024streamingllm,xiao2025duoattention,tang2025razorattention,jiang2024minference,fu2024moa,fu2025headkv,lin2025compresskv}. 
~We further validate this observation through needle-in-a-haystack tests with different context lengths on representative open-source models with diverse architectures and scales, including Mistral-7B-Instruct~\cite{jiang2023mistral}, Qwen3-32B~\cite{yang2025qwen3},  Llama-3.3-70B~\cite{meta2024llama33}, Qwen3.5 397B~\cite{qwen35_397b_a17b_modelcard}  and  DeepSeek V4 Flash~\cite{deepseekv4_flash_modelcard}. As shown in Figure~\ref{fig:global-local-heads}, local heads dominate across all three models: they account for from 83.4\% to 96.8\% of KV heads, while global heads account for 3.2\% to 16.6\%. This result clearly shows that only a small fraction of KV heads are global or prefix-sensitive, while the majority exhibit local and prefix-robust behavior.

This stability allows the system to profile the property of each KV head offline and reuse the profiling result across requests. ~When the same text chunk is reused after a different prefix, the system can therefore focus recovery computation on the small set of global heads that are sensitive to prefix changes, while directly reusing most local-head KV cache. This provides a finer-grained recovery path than token-level recomputation and reduces unnecessary attention computation.
\subsection{Opportunity for Efficient PIC}\label{sec3.3}
The two limitations identified above jointly point to a two-dimensional
opportunity for building an efficient PIC system that preserves output
quality while substantially reducing prefill TTFT.\\
\indent \textbf{\emph{From token-level to head-level recovery.}}~Section~\ref{sec3.1} shows that token-level recomputation cannot exploit
per-head sparsity, because forcing all heads of a selected token to be
recovered together inflates the recomputation set to the union of
head-specific important tokens.
Section~\ref{sec3.2} further shows that only a small fraction of heads
(12.5--15.6\% across representative models) are prefix-sensitive and require
recovery at all, while the vast majority of local heads can be reused
verbatim with negligible quality loss.
These two findings together motivate shifting the recovery granularity from
\emph{tokens} to \emph{heads}: the system recomputes only the global heads
across the entire chunk, leaving local-head KV cache untouched.
Because global/local assignments are stable across requests, they are
determined once via offline profiling and applied at zero per-request
classification overhead.
Head-level recovery thus simultaneously achieves higher fidelity (every
prefix-sensitive head is fully refreshed) and lower cost (no redundant
work on the ${\sim}85\%$ of local heads) than token-level alternatives,
escaping the unfavorable accuracy--latency trade-off described in Section~\ref{sec3.1}.\\
\indent \textbf{\emph{Reducing FFN Cost via Token-Selective Recovery.}}~Section~\ref{sec3.1} also shows that FFN computation contributes significantly to prefill TTFT
across all context lengths, and is the overwhelming bottleneck in short-context, multi-agent workloads where attention-side savings are
inherently limited.
(Section~\ref{sec:algorithmic_sparsity})~shows that FFN computation contributes significantly to prefill TTFT. RedKnot reduces this cost through token-selective FFN recovery. After head-aware attention recovery, RedKnot estimates token importance from the recovered attention signal. Important token states execute the original dense FFN, while other token states bypass the FFN update through the residual path. This optimization is independent of KV-cache sparsity and can accelerate short-context workloads where attention-side savings are limited.
This sparsity is structurally independent of the KV cache and can be
exploited in parallel with head-level KV reuse.
By evaluating only the predicted active neurons per token, the system skips
the majority of FFN multiply-accumulate operations without altering the
attention computation path.
Critically, FFN sparsity is effective regardless of context length, making
it the primary lever for accelerating short-context scenarios where KV reuse
alone cannot deliver meaningful TTFT reduction.
\subsection{Challenges}\label{sec3.4}
Although the sparsity identified in Sections~\ref{sec3.1} and~\ref{sec3.2}
provides a clear opportunity, translating it into real TTFT reduction,
lower computational cost, and high-accuracy outputs raises two practical
challenges.\\
\noindent\textbf{C1: Discrete per-head sparsity undermines token-level recovery.}
~Although attention sparsity is widely recognized~\cite{xiao2025duoattention,tang2025razorattention,jiang2024minference}, the discrete and heterogeneous nature of per-head sparsity makes it fundamentally difficult to exploit at the token level.~Under multi-head attention, different heads attend to different subsets of tokens, and their sparse patterns vary substantially across heads.~When a token-level PIC system selects a fixed set of tokens for recomputation, it must take the union of the important token sets across all heads in order to satisfy every head.~This union quickly expands to cover a large fraction of the chunk, forcing the system to recompute nearly as many tokens as a full prefill, which defeats the purpose of KV cache reuse.
~Conversely, if the system selects only a small token subset to limit recomputation cost, a different failure mode arises.~For any head whose important tokens are \emph{not} included in the selected set, the attention scores computed during correction will be based on an incomplete context, leaving the error in that head's KV cache unaddressed and degrading output quality.~Furthermore, even for tokens that \emph{are} selected for recomputation, the correction remains incomplete.
~Consider token $i$ chosen as an important token shared across all heads: because tokens $0$ to $i-1$ still use the stale reused KV cache rather than freshly computed values, the attention context seen by token $i$ during recomputation is itself corrupted.~As a result, the recomputed KV of token $i$ cannot fully recover to its ground-truth full-prefill value, and this error propagates forward through the sequence.~This cascading inaccuracy is an inherent limitation of any token-level PIC scheme: reusing KV entries for preceding tokens while selectively recomputing later ones introduces an irrecoverable dependency on stale context.\\
\noindent \textbf{C2: Long contexts amplify noise and dilute critical evidence.}
Long-context execution can degrade model quality, not only efficiency. Although modern LLMs advertise increasingly long context windows, a growing body of evidence shows that their effective context utilization remains far below the nominal window size. Models often fail to reliably use relevant evidence when it is placed in hard-to-access positions, mixed with large amounts of background text, or embedded in long multi-document inputs~\cite{liu2024lost,hsieh2024ruler,bai2024longbench,bai2025longbenchv2,li2024needlebench,modarressi2025nolima}. More seriously, recent studies show that performance can degrade as the input length increases even when the required evidence is available or retrieval is controlled, suggesting that long input length itself can hurt reasoning and answer quality~\cite{du2025contextlength,wu2024longgenbench}.\\
\indent This problem is particularly severe in long-context RAG and agent workloads. A prompt may contain retrieved passages, tool outputs, execution traces, memory snapshots, and historical dialogue states. However, only a small fraction of these tokens are directly useful for the current query. The rest may be weakly related, redundant, stale, or irrelevant. As the context grows from thousands to hundreds of thousands of tokens, useful evidence becomes increasingly sparse relative to the background context. Dense prefill forces every token to participate in attention and every token state to pass through FFN layers, allowing irrelevant or low-value tokens to consume computation, perturb hidden states, and propagate noisy updates through the residual stream.~Therefore, the challenge is not merely that long contexts are expensive to process; they also make useful information harder to preserve. Critical evidence tokens must compete with a much larger set of background tokens across attention and FFN computation. This dilutes their salience, increases the risk of position-dependent retrieval failure, and places a heavier burden on later layers to recover the correct signal. A long-context serving system must therefore account for the signal-to-noise imbalance introduced by dense execution over large prompts, rather than treating all heads and all tokens as equally useful.\\
\noindent\textbf{C3: Token-level KV layout amplifies HBM traffic.}
~PIC(Position-Independent Cache) is primarily motivated by long-context RAG, where retrieved passages are
concatenated into prompts spanning tens to hundreds of thousands of tokens. ~In
this setting, KV-cache traffic becomes a dominant cost because the cache size
scales linearly with sequence length and proportionally with the number of
layers and attention heads~\cite{pope2023efficiently}. ~For a 70B-class model
at 128K context length, the KV cache alone can exceed 40,GB, consuming a large
fraction of GPU HBM~\cite{agrawal2024sarathi}. ~Modern GPUs make this bottleneck
more pronounced: compute throughput has grown much faster than HBM bandwidth.
The compute--memory balance point is already about ${\sim}156$,FLOP/Byte on
A100 and ${\sim}295$,FLOP/Byte on H100 SXM~\cite{nvidia2023h100,
choquette2023nvidia}. Long-context LLM inference, especially decode and sparse
prefill, often falls below this threshold and is therefore limited by memory
traffic rather than arithmetic throughput~\cite{pope2023efficiently,
kwon2023vllm,agrawal2024sarathi,zhong2024distserve}.\\
\indent This bandwidth bottleneck is further amplified when PIC applies different reuse and recovery policies across heads. In head-aware PIC, global heads may require full-context recovery or long-range KV access, whereas local heads only need sink tokens and a short recent window. This creates inherently heterogeneous access patterns across heads, including different read, write, transfer, and eviction frequencies. However, existing KV-cache managers are typically organized at token-block granularity, where each block stores the KV states of multiple heads over the same token range. Consequently, accessing or updating the KV state of one head may implicitly load, move, or rewrite unrelated heads in the same block, causing read/write amplification and wasting HBM bandwidth.~This layout mismatch also limits distributed and disaggregated serving. Under prefill--decode disaggregation or remote KV reuse, the system should ideally transfer only the head-specific states consumed by the decode worker. A token-level KV layout instead couples all heads within the same token range, forcing dense block transfer even when most heads are local, stale, or unnecessary. Therefore, efficient PIC requires a head-addressable KV layout that exposes head-level sparsity to memory management, data movement, and attention execution.\\

\vspace{-1.5ex}
\section{\sysname\ Design}\label{sec4}
\begin{figure*}[t]
  \centering
  \includegraphics[width=0.9\textwidth]{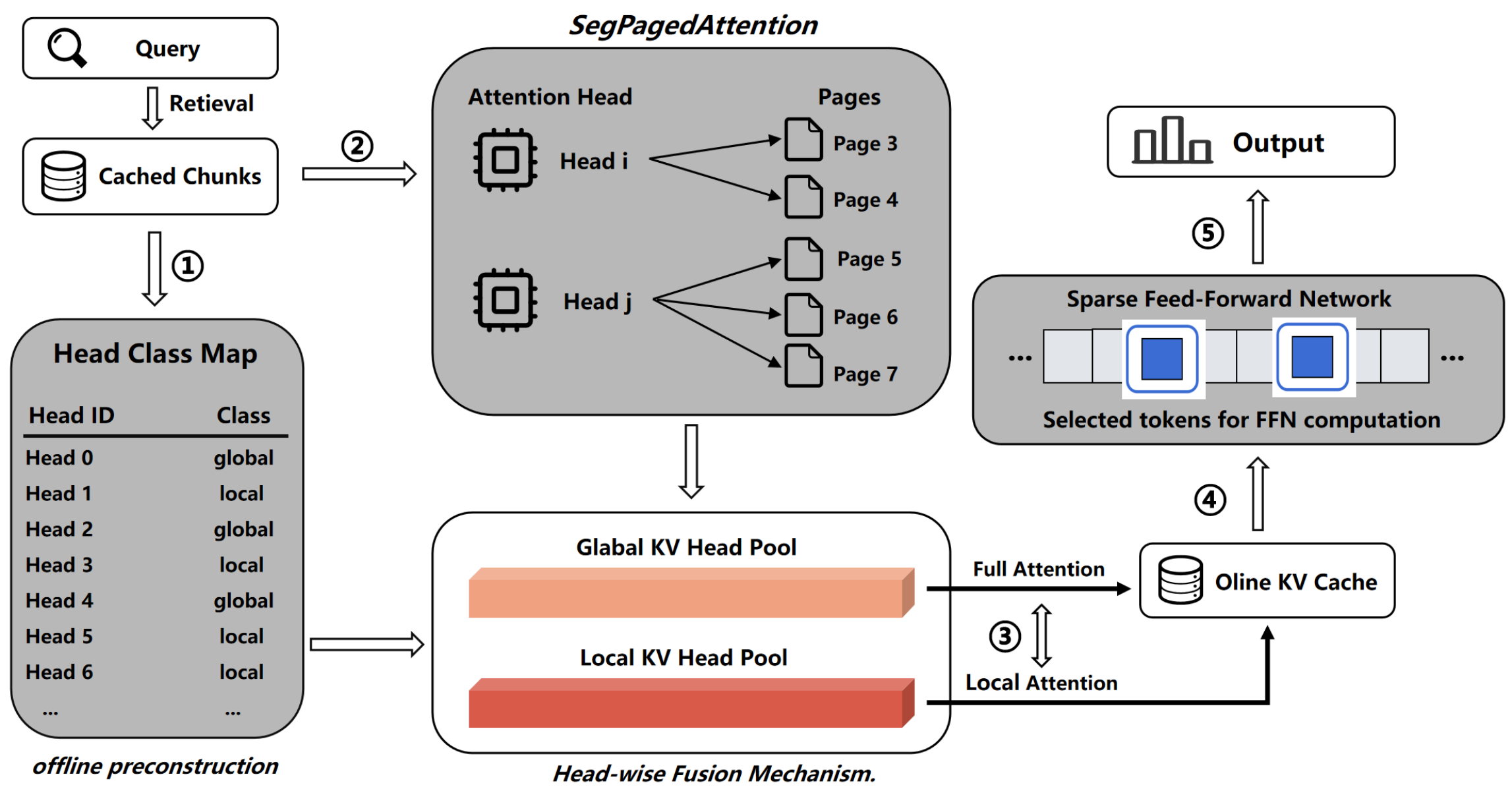}
  \caption{Overview of \sysname.}
  \vspace{-3ex}
  \label{fig:OverviewofRedKnot}
\end{figure*}
\vspace{-1.5ex}
In this section, we introduce \sysname, a system that accelerates long-context
LLM inference by addressing the challenges discussed in Section~\ref{sec3.4} with two key components: (i) \method, which
exploits head-level KV and FFN sparsity to reduce redundant computation while preserving high output accuracy; and (ii) \SegPagedAttention, a head-aware KV-cache layout
that avoids loading irrelevant head data and improves HBM bandwidth utilization and throughput.
\subsection{Overview of \sysname}
\sysname\ consists of two core components: (i) an Elastic Sparsity module, and (ii) a module that stores data at the granularity of KV-cache heads.
~We next describe the end-to-end workflow of \sysname, highlighting how these modules interact during inference.
~As shown in Figure~\ref{fig:OverviewofRedKnot}, ~\ding{182}~During the offline stage, we run inference on the text to generate the KV cache. Meanwhile, through profiling, we construct for each model a hashmap of KV-cache heads indexed at the \((\text{layer}, \text{head})\) granularity, thereby establishing a mapping table that records the global and local attributes of each KV-cache head.  ~\ding{183}Next, we store the KV cache at the granularity of KV-cache heads and establish a mapping between the segments of each KV-cache head and their corresponding virtual page numbers. ~\ding{184} When an online query arrives, we load the relevant text and fully recompute the global KV-cache heads, except when the corresponding KV cache lies in the prefix. For local KV-cache heads, we largely reuse the cached values and recompute only a small portion of them.  ~\ding{185} 
After the KV cache of a layer has been computed, we adopt a partially sparse FFN strategy to reduce computational overhead while mitigating attention noise from the KV cache in deeper layers.~\ding{186}~\sysname~enables efficient and systems-friendly long-context LLM serving with low compute cost, low TTFT, and high accuracy. ~Next, We present each component in detail in the following sections.
\subsection{\method}
\begin{figure}[t]
  \centering
\includegraphics[width=0.98\linewidth]{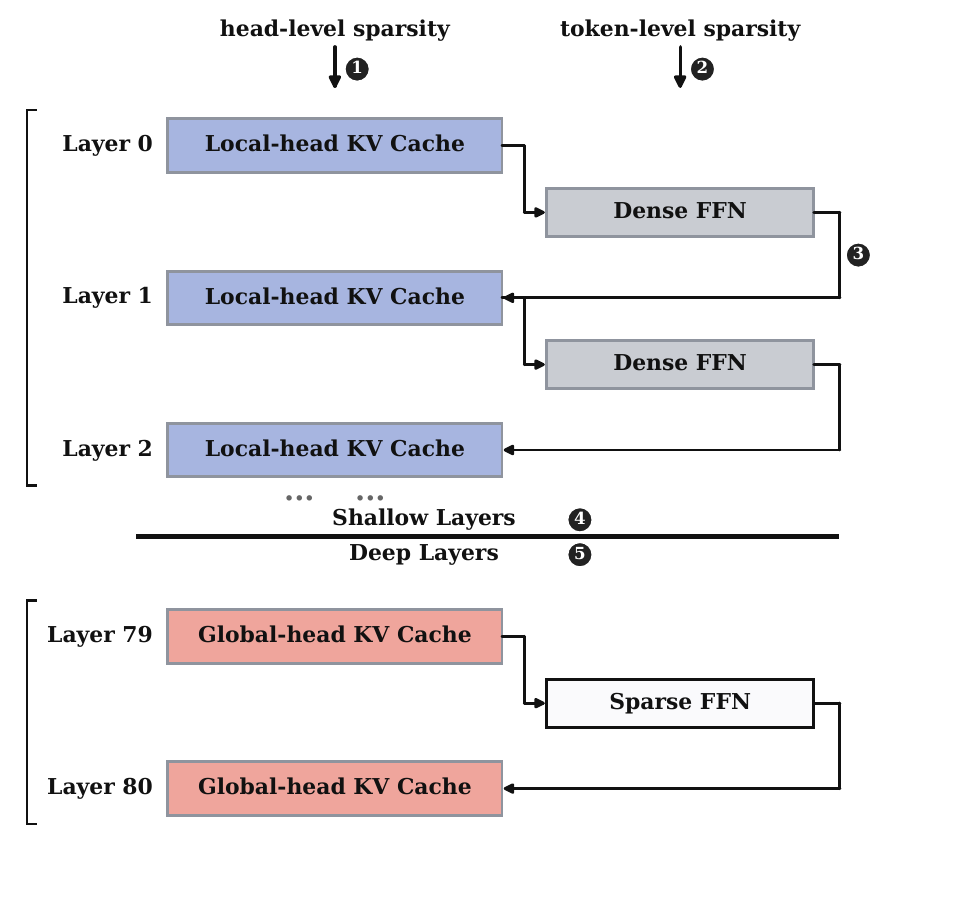}
  \caption{WorkFlow of Elastic Sparsity.}
  \vspace{-3ex}
  \label{fig:method-overview}
\end{figure}

To address \emph{Challenge 1 and Challenge 2} discussed in Section~\ref{sec3.4}, we design an algorithm named \emph{Elastic Sparsity} to achieve high accuracy. As shown in ~Figure~\ref{fig:method-overview}, the overall algorithmic workflow of \method.~\ding{182} For the KV cache,~\method~applies a multi-head sparse policy, exploiting head-level sparsity across attention heads.~\ding{183} For the FFN, ~\method~ applies token-level sparsity: important tokens are selected to execute the full FFN.~\ding{184} Unimportant tokens directly follow the residual connection, which reduces computation while making important tokens more salient in the recovered representations.~\ding{185} In shallow layers, ~\method~reuses most local-head KV states and keeps FFN computation dense to preserve the early residual stream.~\ding{186} In deep layers, ~\method~recomputes prefix-sensitive global-head KV states and applies sparse FFN computation only to important tokens.

The goal of \method\ is to recover the quality of position-independent KV cache reuse without replaying the full dense prefill path. \method\ decomposes recovery into three steps: ~RoPE-based positional alignment, ~head-aware attention recovery, and ~partial sparse FFN recovery.\\
\indent \textbf{\emph{RoPE-based positional alignment.}}~As a dominant positional encoding scheme in modern LLMs, ~RoPE exhibits rotation invariance in the sense that positional shifts can be represented as relative rotations in the query--key inner product~\cite{su2024roformer,chen2023positioninterpolation,liu2025rethinkingrope}.~When a reusable chunk is cached offline and later placed after a different prefix, its token positions change. Since modern LLMs commonly use RoPE, the cached keys contain position-dependent rotations. Before applying recovery, \method\ first aligns the cached keys to their online positions using the rotational structure of RoPE. For a cached key originally encoded at offline position $p_{\mathrm{off}}$ and reused at online position $p_{\mathrm{on}}$, \method\ applies the relative rotation
\[
K(p_{\mathrm{on}}) = R(p_{\mathrm{on}})R(p_{\mathrm{off}})^{-1}K(p_{\mathrm{off}}),
\]
where $R(\cdot)$ denotes the RoPE rotation matrix. This step removes the deterministic position mismatch caused by moving the chunk to a new location. The remaining error mainly comes from contextual mismatch, i.e., the fact that the chunk now appears after a different prefix. \method\ then applies head-aware recovery to correct this contextual mismatch.\\
\indent \textbf{\emph{Layer-wise elastic recovery.}}
\method\ adopts a layer-wise recovery strategy. In shallow layers, \method\ uses local attention recovery with full FFN computation. This conservative design preserves the early residual stream, where hidden states are more sensitive to perturbations and errors can be amplified by subsequent layers. In deep layers, \method\ enables global-head attention recovery together with sparse FFN computation. Since deeper layers exhibit stronger semantic selectivity and more concentrated attention behavior, \method\ recomputes prefix-sensitive global heads, reuses prefix-robust local heads, and applies FFN only to important tokens.\\
\begin{algorithm}[t]
\caption{\method\ Elastic Sparse Recovery}
\label{alg:redknot-elastic}
\begin{algorithmic}[1]
\Require New prefix $P$, reusable chunk $C$, cached KV states $\mathcal{K}_{C}$
\Require Head class map $\mathcal{M}$, local window size $w$, sink set $\mathcal{S}_{\mathrm{sink}}$, dense-layer boundary $L_{\mathrm{dense}}$
\Ensure Recovered hidden states for $P \Vert C$
\State $\widetilde{\mathcal{K}}_{C} \leftarrow \mathrm{RoPEAlign}(\mathcal{K}_{C})$
\Comment{align cached keys to online positions}
\State $X_0 \leftarrow \mathrm{Embed}(P \Vert C)$
\For{$\ell = 0$ to $L-1$}
    \State $\mathcal{H}_{g} \leftarrow \{h \mid \mathcal{M}(\ell,h)=\textsc{Global}\}$
    \State $\mathcal{H}_{l} \leftarrow \{h \mid \mathcal{M}(\ell,h)=\textsc{Local}\}$

    \If{$\ell < L_{\mathrm{dense}}$}
        \State $A_{\ell} \leftarrow \mathrm{LocalAttention}(X_{\ell}, \widetilde{\mathcal{K}}_{C}, \mathcal{H}_{l})$
        \State $Y_{\ell} \leftarrow X_{\ell} + A_{\ell}$
        \State $X_{\ell+1} \leftarrow Y_{\ell} + \mathrm{FFN}_{\ell}(Y_{\ell})$
        \Comment{full FFN in shallow layers}
    \Else
        \State $A_{\ell}^{g} \leftarrow \mathrm{RecomputeGlobalHeads}(X_{\ell}, P, C, \mathcal{H}_{g})$
        \For{local head $h \in \mathcal{H}_{l}$}
            \For{token $i \in C$}
                \State $\mathcal{W}(i) \leftarrow \mathcal{S}_{\mathrm{sink}} \cup [\max(0,i-w), i]$
                \State $A_{\ell,h,i} \leftarrow \mathrm{RepairLocalHead}(X_{\ell}, \widetilde{\mathcal{K}}_{C,h}, \mathcal{W}(i))$
            \EndFor
        \EndFor
        \State $A_{\ell} \leftarrow \mathrm{MergeHeads}(A_{\ell}^{g}, A_{\ell}^{l})$
        \State $Y_{\ell} \leftarrow X_{\ell} + A_{\ell}$
        \State $S_{\ell} \leftarrow \mathrm{SelectImportantTokens}(A_{\ell})$
        \State $Z_{\ell}[S_{\ell}] \leftarrow \mathrm{FFN}_{\ell}(Y_{\ell}[S_{\ell}])$
        \State $Z_{\ell}[\overline{S_{\ell}}] \leftarrow 0$
        \Comment{residual identity}
        \State $X_{\ell+1} \leftarrow Y_{\ell} + Z_{\ell}$
    \EndIf
\EndFor
\State \Return $X_L$
\end{algorithmic}
\end{algorithm}
\begin{figure*}[t]
  \centering
\includegraphics[width=0.8\textwidth,height=0.25\textheight]{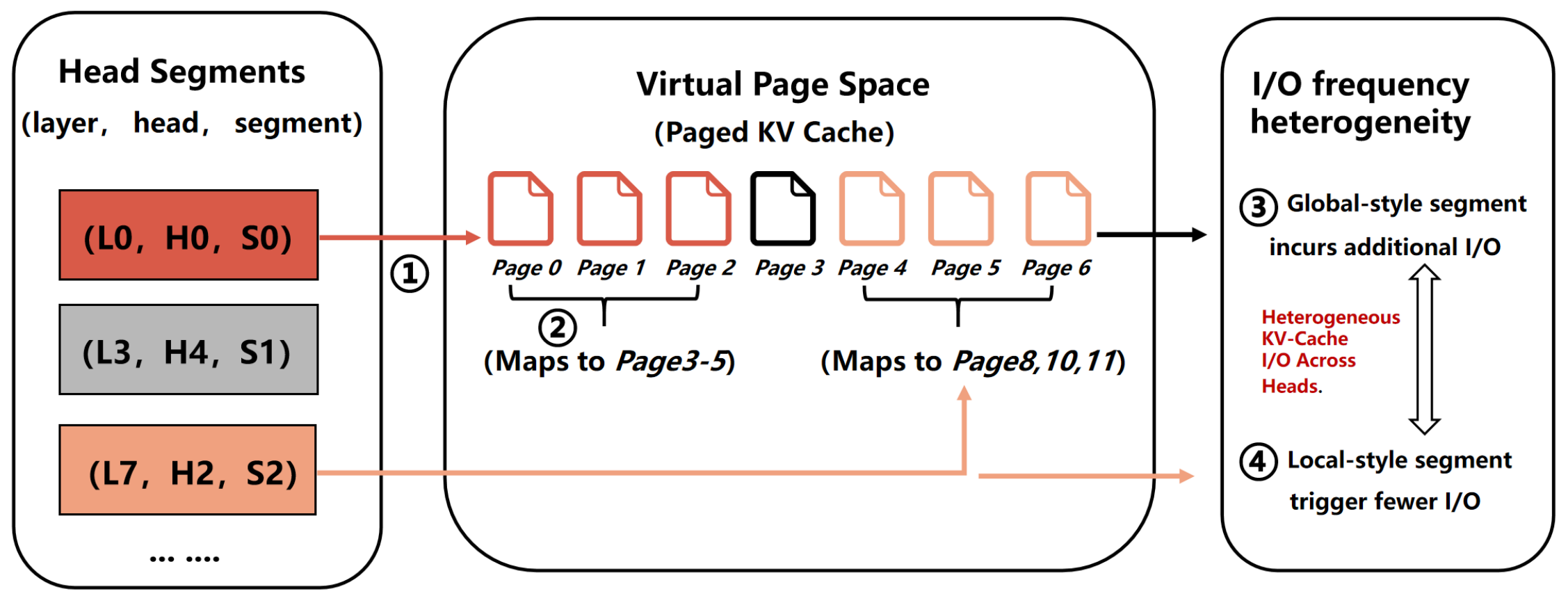}
  \caption{Overview of \SegPagedAttention.}
  \vspace{-3ex}
  \label{fig:SegPagedAttention}
\end{figure*}
%
\indent \textbf{\emph{Head-aware attention recovery.}}
For global heads, \method\ recomputes their KV states under the new prefix, unless the reused chunk is already covered by standard prefix reuse. This conservative policy preserves long-range dependency modeling and prevents prefix-induced errors from propagating through retrieval-oriented heads. For local heads, \method\ reuses most cached KV states because their effective attention range is bounded. Specifically, for a local head at token $i$, \method\ only recomputes the local visible set
\[
\mathcal{W}(i)=\mathcal{S}_{\mathrm{sink}} \cup [\max(0,i-w), i],
\]
where $w$ is the local window size and $\mathcal{S}*{\mathrm{sink}}$ denotes the reserved sink-token positions. For example, when $i=1000$ and $w=256$, \method\ recomputes the sink tokens and the local window around positions $744$ to $1000$, while directly reusing the remaining local-head KV cache. This policy preserves local continuity while avoiding unnecessary recomputation over distant tokens that are invisible to local heads.\\
\indent \textbf{\emph{Partial sparse FFN recovery.}}
Head-aware attention recovery reduces attention-side recovery cost, but FFN computation can still remain on the critical path of prefill. Therefore, after the recovered attention states are produced in deep layers, \method\ applies partial sparse FFN recovery. \method\ uses the recovered attention signal to estimate token importance. Tokens with high importance execute the dense FFN, while other tokens follow the residual identity path. In this way, \method\ spends FFN computation only where correction is likely to affect the final hidden states.\\
\indent The two sparsity dimensions are complementary. Head-aware attention recovery reduces unnecessary KV recomputation across heads, while partial sparse FFN recovery avoids dense FFN replay across tokens. Together, they form an elastic sparsity mechanism that adapts recovery cost to the correction demand of each layer and reused chunk.
Algorithm~\ref{alg:redknot-elastic} summarizes the elastic sparse recovery procedure in \method. Lines~1--2 first initialize the online recovery state. Specifically, \method\ applies \textsc{RoPEAlign} to rotate cached keys from their offline positions to the online positions in the composed prompt, and then initializes the hidden states for the new prefix and reusable chunk. Lines~3--5 iterate over Transformer layers and query the offline head class map $\mathcal{M}$ to obtain the global heads $\mathcal{H}_g$ and local heads $\mathcal{H}_l$ for each layer.~Lines~6--9 handle shallow layers. In these layers, \method\ uses local attention recovery while keeping the FFN computation dense. This conservative policy preserves the early residual stream and avoids amplifying errors in subsequent layers. Lines~10--23 handle deep layers, where \method\ enables elastic sparse recovery. Lines~11--17 perform head-aware attention recovery: global heads are recomputed under the new prefix, while local heads are repaired only within their visible local window $\mathcal{W}(i)=\mathcal{S}_{\mathrm{sink}}\cup[\max(0,i-w),i]$. This allows \method\ to correct prefix-sensitive heads while reusing most prefix-robust local-head KV states.
After attention recovery, Lines~18--19 merge the recovered heads and form the intermediate hidden states. Lines~20--23 then apply partial sparse FFN recovery. \method\ selects important tokens according to the recovered attention signal, executes the dense FFN only on these selected tokens, and sets the FFN update of unselected tokens to zero so that they follow the residual identity path. Finally, the algorithm returns the recovered hidden states. Overall, Algorithm~\ref{alg:redknot-elastic} combines RoPE-based positional alignment, head-aware attention recovery, and sparse FFN recovery to reduce PIC recovery cost while preserving output fidelity.
\subsection{\SegPagedAttention}
To address \emph{Challenge 3} discussed in Section ~\ref{sec3.4}, we design an
module named \SegPagedAttention to achieve high concurrency.~Beyond this, \SegPagedAttention ~also provides a general system substrate for existing head-sparse KV-cache algorithms discussed in Section~\ref{sec:algorithmic_sparsity}, which otherwise have to \textbf{express head-level sparsity on top of conventional token-block-based PagedAttention.} ~\SegPagedAttention~is the storage and execution substrate that supports \method's head-aware recovery policy. ~Existing paged KV-cache systems usually organize cached states at token-block granularity, where all KV heads within the same token block are allocated, transferred, and accessed together. ~This abstraction is efficient for conventional prefix caching, but it is too coarse-grained for head-aware PIC recovery. In \method, different heads may follow different runtime policies: global-style heads always require full-range attention over all mapped KV pages, while local-style heads mostly need only sink tokens and recent-window pages. ~Therefore, the KV cache should expose the head dimension to the runtime. ~To this end, \SegPagedAttention\ introduces a head-segmented KV layout. As shown in Figure~\ref{fig:SegPagedAttention}, ~\ding{182}instead of treating a token block as the only management unit, \SegPagedAttention\ represents cached states as \emph{head segments}. ~Each head segment is indexed by $(\ell,h,s)$, where $\ell$ is the layer id, $h$ is the KV head id, and $s$ is the segment id within that head stream. ~\ding{183} A head segment contains a contiguous range of KV states for a single KV head. To remain compatible with existing PagedAttention allocators and fragmented physical memory, \SegPagedAttention introduces a virtual-page indirection: each head segment is mapped to one or more consecutive virtual pages, while these virtual pages are further mapped to non-contiguous physical KV pages managed by the underlying PagedAttention backend. This design preserves the existing paged KV-cache infrastructure while exposing a head-addressable logical layout to the runtime. ~\ding{184} Storage pages backing global-style segments exhibit higher I/O access frequency, because global heads require full-context attention or long-range KV recovery.~\ding{185} In contrast, storage pages backing local-style segments usually exhibit much lower I/O access frequency in most KV-cache reuse scenarios, since local heads only consume sink tokens and a short recent window. This head-level disaggregated layout substantially reduces I/O amplification by avoiding unnecessary accesses to unrelated heads, thereby improving decode efficiency and serving throughput.\\
\indent The key advantage of this layout is that \textbf{it decouples head-specific execution policies from token-block storage}. ~With token-level paging, if any head inside a token block requires full-context recovery, the system tends to load or recompute the entire block across all heads. ~This collapses head-specific sparsity into a coarse token-level decision. ~In contrast, \textbf{SegPagedAttention allows the runtime to access only the pages associated with the required head segment}. ~A global-style segment can access its full mapped page range and execute full-range attention except when it happens to be in the prefix, while a local-style segment can access only the pages that overlap with its visible region, such as sink pages and recent-window pages. ~Thus, different head segments can follow different attention execution policies without forcing all heads in the same token range to be processed together. ~\SegPagedAttention\ is especially useful for position-independent KV reuse. ~When a cached chunk is reused after a new prefix, \method\ recomputes or repairs only the prefix-sensitive head segments, while directly reusing the prefix-robust local segments whenever their visible pages are unchanged. ~This avoids the token-level union problem discussed in Section~\ref{sec3.1}: ~even if different heads require different correction regions, their sparse structures can be preserved separately at the head-segment level. ~As a result, \SegPagedAttention\ makes KV cache both pageable and head-aware, providing the system support required by elastic sparse recovery. \\
\indent The main execution flow is summarized in Algorithm~\ref{alg:segpagedattention}. Lines~1--14 construct the head-specific metadata used by \SegPagedAttention, while Lines~15--17 perform the actual fused attention execution. Specifically, Line~1 initializes an empty metadata buffer. Lines~2--5 iterate over KV heads in the current layer, collect all logical head segments belonging to each head, query the segment table, and assemble the corresponding virtual page list. Lines~6--11 then apply the head execution policy: global-style heads keep the full virtual page range, whereas local-style heads retain only sink pages and recent-window pages. Lines~12--13 translate the selected virtual pages into physical backing pages through the underlying paged allocator and pack the result into the per-layer varlen metadata. After all heads have been processed, Line~15 invokes a fused varlen attention kernel over the packed head-specific page lists, avoiding independent attention launches for each token block or segment. Finally, Lines~16--17 merge the per-head outputs and return the attention result for the current layer. This design preserves paged KV-cache management while exposing head segments as independent execution units, thereby avoiding dense token-block accesses to unrelated heads.
\begin{algorithm}[t]
\caption{SegPagedAttention}
\label{alg:segpagedattention}
\begin{algorithmic}[1]
\Require Query states $Q_{\ell}$, KV heads $\mathcal{H}*{\ell}$
\Require Segment table $\mathcal{T}*{\mathrm{seg}}$, physical page table $\mathcal{T}*{\mathrm{page}}$
\Require Head policy map $\mathcal{P}$, sink pages $\mathcal{S}*{\mathrm{sink}}$, local window size $w$
\Ensure Attention output $A_{\ell}$ for layer $\ell$
\State Initialize segment metadata $\mathcal{M}_{\ell} \leftarrow \emptyset$
\For{each KV head $h \in \mathcal{H}*{\ell}$}
\State $\mathcal{G}*{\ell,h} \leftarrow \mathrm{HeadSegments}(\ell,h)$
\Comment{segments belonging to head $h$}
\State $\mathcal{V}*{h} \leftarrow \bigcup*{g \in \mathcal{G}*{\ell,h}} \mathcal{T}*{\mathrm{seg}}[g]$
\Comment{logical virtual pages}
\State $p \leftarrow \mathcal{P}(\ell,h)$
\Comment{head execution policy}

\If{$p=\textsc{Global}$}
    \State $\mathcal{U}_{h} \leftarrow \mathcal{V}_{h}$
    \Comment{full visible range}
\ElsIf{$p=\textsc{Local}$}
    \State $\mathcal{R}_{h} \leftarrow \mathrm{RecentPages}(\mathcal{V}_{h}, w)$
    \State $\mathcal{U}_{h} \leftarrow \mathcal{S}_{\mathrm{sink}} \cup \mathcal{R}_{h}$
    \Comment{sink and recent-window pages}
\EndIf

\State $\mathcal{B}_{h} \leftarrow \mathrm{Translate}(\mathcal{T}_{\mathrm{page}}, \mathcal{U}_{h})$
\Comment{physical backing pages}
\State $\mathcal{M}_{\ell} \leftarrow \mathcal{M}_{\ell} \cup \{(h,\mathcal{B}_{h},|\mathcal{U}_{h}|,p)\}$
\EndFor

\State $\mathcal{O}*{\ell} \leftarrow \mathrm{FusedVarlenAttention}(Q*{\ell}, \mathcal{M}*{\ell})$
\State $A*{\ell} \leftarrow \mathrm{MergeHeads}(\mathcal{O}*{\ell})$
\State \Return $A*{\ell}$
\end{algorithmic}
\end{algorithm}
\subsection{Architecture-Agnostic Implementation}
\label{sec:architecture_agnostic_impl}
\sysname~is implemented as an architecture-agnostic runtime rather than a model-specific sparse-attention kernel. The key abstraction is a \emph{reusable state object}. Each state object is associated with a layer, a head or head group, a segment range, a state type, a position transform, and an execution policy. ~As disgussed in section~\ref{sec:hybirdattention&MLA}~depending on the model architecture, the state object may correspond to explicit KV pages, MLA latent states, or recurrent linear-attention states. This abstraction allows \sysname to apply the same high-level policies---offline state construction, head-level classification, position-aware recovery, and token-level FFN sparsity---across different long-context model families.\\
\indent \textbf{\emph{Unified adapter interface.}}
To support heterogeneous architectures, \sysname separates policy decisions from model-specific state execution. Each architecture backend implements four adapter functions. First, \textsc{Profile} measures the effective attention or memory range of each layer and head, and produces a head policy map. Second, \textsc{BuildState} constructs reusable offline states for cached chunks. Third, \textsc{SelectVisibleState} determines which state segments are visible to each head at runtime, such as full-context pages, sink pages, recent-window pages, compressed latent pages, or recurrent prefix states. Finally, \textsc{Execute} invokes the corresponding backend operator, such as fused attention over explicit KV pages, FlashMLA over latent states, or recurrent state composition for linear attention. Therefore, the scheduler and cache manager operate over a common state interface, while architecture-specific adapters handle how the state is materialized and consumed.\\
\indent \textbf{\emph{Adapting Qwen3.5-style hybrid models.}}
Qwen3.5-style models combine full-attention layers, Gated DeltaNet linear-attention layers, and sparse MoE feed-forward blocks~\cite{qwen35_35b_a3b_modelcard,qwen35_397b_a17b_modelcard,yang2025gateddeltanet}. This architecture exposes two different forms of reusable history. Full-attention layers still produce explicit KV cache, and thus can be handled by the standard head-aware KV path in \sysname. For these layers, \sysname applies conservative dense execution in shallow layers and head-class recovery in deeper layers: global heads preserve full-context access, while local heads reuse sink and recent-window KV pages.~In contrast, Gated DeltaNet layers do not expose a full token-level KV cache, but they still maintain \emph{multi-head} recurrent states. Therefore, \sysname applies the same head-aware principle to linear-attention layers. During offline profiling, \sysname measures the memory behavior of each $(\ell,h)$ pair and classifies linear-attention heads into global and local classes. For global linear heads, whose recurrent states depend on long-range history, \sysname recomputes the corresponding recurrent states under the newly composed prefix and reused chunks, so that prefix-conditioned cross-segment dependencies are correctly reflected. For local linear heads, \sysname estimates an effective window size $w_{\ell,h}$ and caches the out-of-window recurrent state as a compact prefix-state checkpoint. When computing token $i$, the adapter initializes the recurrence from the cached state before the visible window and only replays the state updates within the recent window $[i-w_{\ell,h}, i]$.\\
\indent This design preserves the multi-head structure of linear attention while avoiding the materialization of all previous tokens. Global linear heads retain full-history correctness through recomputation, whereas local linear heads reuse cached prefix-state checkpoints and only refresh their visible windows. For MoE blocks, \sysname uses full-attention layers as token-importance synchronization points. Specifically, a full-attention layer computes attention mass over tokens and selects a salient-token set, such as the smallest set of tokens whose cumulative attention mass exceeds a threshold $\tau$. The following linear-attention layers reuse this salient-token set for token-level sparse FFN execution until the next full-attention layer refreshes it. Selected tokens execute the full MoE path, while low-importance tokens follow a lightweight path, such as residual identity or a downgraded expert path. As a result, the Qwen3.5 backend combines explicit-KV reuse for full-attention layers, head-wise recurrent-state reuse for linear-attention layers, and token-level MoE sparsity under the same runtime policy.\\
\indent \textbf{\emph{Adapting DeepSeek-V4-style MLA models.}}
DeepSeek-V4-style models use MLA or MLA-derived compressed attention states instead of ordinary per-head KV tensors. In these models, the physical cache is a packed latent KV stream, while logical attention heads reconstruct their head-specific keys and values from the shared latent state. Directly expanding the latent cache into explicit per-head KV pages during serving would destroy the memory advantage of MLA. Therefore, \sysname only uses a decompressed head-wise view for offline profiling and keeps the runtime cache in the native MLA representation.~\sysname~adapts MLA with an offline--online aggregation path. During offline profiling, \sysname~analyzes the decompressed MLA-induced head-wise KV states and classifies logical heads into global and local heads. Global heads are treated as prefix-sensitive heads and are not approximated by offline compression. Instead, their KV states are recomputed online under the newly composed prefix. For local heads, \sysname estimates an effective window size and stores only the out-of-window history as a compressed offline MLA state, denoted as ~$\mathbf{MLA}_{\mathbf{offline}}$. This offline state summarizes distant history that is outside the local visible window.\\
\indent~During online serving, \sysname~recomputes the MLA states needed by the current request, including the new prefix, global-head KV states, and the local-head visible window.~Since most local heads in DeepSeek-V4 operate with a 128-token sliding attention window, \sysname~\textbf{recomputes the first 128 tokens of each non-initial chunks}. These boundary tokens are most affected by the missing cross-chunk context after position-independent reuse, and therefore incur the largest contextual mismatch. In contrast, for tokens from position 128 to the end of chunk (i), the attention mass of local heads is almost entirely concentrated within the chunk itself, allowing their cached MLA states to be reused with minimal fidelity loss.~These online states form ~$\mathbf{MLA}_{\mathbf{online}}$. The RoPE-related component is handled as architecture-specific position metadata and is realigned to the current online positions before fusion. Finally, \sysname ~aggregates $\mathrm{MLA}*{\mathrm{offline}}$ and $\mathrm{MLA}*{\mathrm{online}}$ through log-sum-exp softmax fusion, producing the materialized MLA attention state used by subsequent computation. This aggregation is exact with respect to the materialized offline and online states: splitting the key set into offline and online parts and merging them with LSE gives the same result as applying softmax attention over their union. When offline compression is disabled, this fused MLA path is numerically equivalent to dense full attention; with compression enabled, the approximation comes only from the compressed $\mathrm{MLA}*{\mathrm{offline}}$ state. ~MoE computation follows the same layer-wise sparse FFN policy as Elastic Sparsity. Shallow layers keep dense MoE execution to preserve the early residual stream, while deeper layers apply token-level sparsity: important tokens execute the full MoE block, and low-importance tokens follow the residual path or a lightweight expert path. ~We leverage the built-in \textbf{indexer signal} of the DeepSeek-V4 framework as the sparsity indicator, since it directly reflects the model’s native sparse structure. We then select the top-ranked tokens according to this signal, which helps suppress noisy or low-importance tokens in long-context scenarios.~Thus, the DeepSeek-V4 backend combines global-head online recomputation, local-head offline MLA compression, online MLA recomputation, LSE-based MLA aggregation, and layer-wise sparse MoE execution within the same architecture-agnostic runtime interface.\\
\indent Because the Qwen 3.5 series and DeepSeek V4 models differ significantly from standard GQA models, SegPagedAttention cannot be adapted simply on a per-head basis. Since the Qwen 3.5 series and DeepSeek V4 can obtain the sparse features of the next layer’s KV from layer i—for example, DeepSeek V4’s indexer signal and the key head and token information identified in Qwen 3.5 offline testing—\textbf{we align SegPaged Attention with the indexer-selected KV in DeepSeek V4 and the sparse KV in Qwen 3.5, temporarily reordering them so that the discrete sparse tokens are prefetched into segments before they are used}.

%
\section{Evaluation}
\label{sec:eval}
We evaluate \sysname\ along two axes: (1)~\emph{accuracy evaluation},
measured by standard QA metrics (F1, EM) and logit-level fidelity
indicators (cosine similarity, top-1/top-10 agreement), and
(2)~\emph{system efficiency}, measured by time-to-first-token (TTFT), FLOPs
breakdown, KV-cache bandwidth, and throughput.
All experiments run on a single 8-GPU node; the same physical hardware
serves both the standalone prefill benchmarks and the prefill--decode
(PD) disaggregation experiments.
\subsection{Experimental Setup}
\label{sec:eval:setup}
\begin{figure*}
  \centering
  \includegraphics[width=0.98\textwidth]{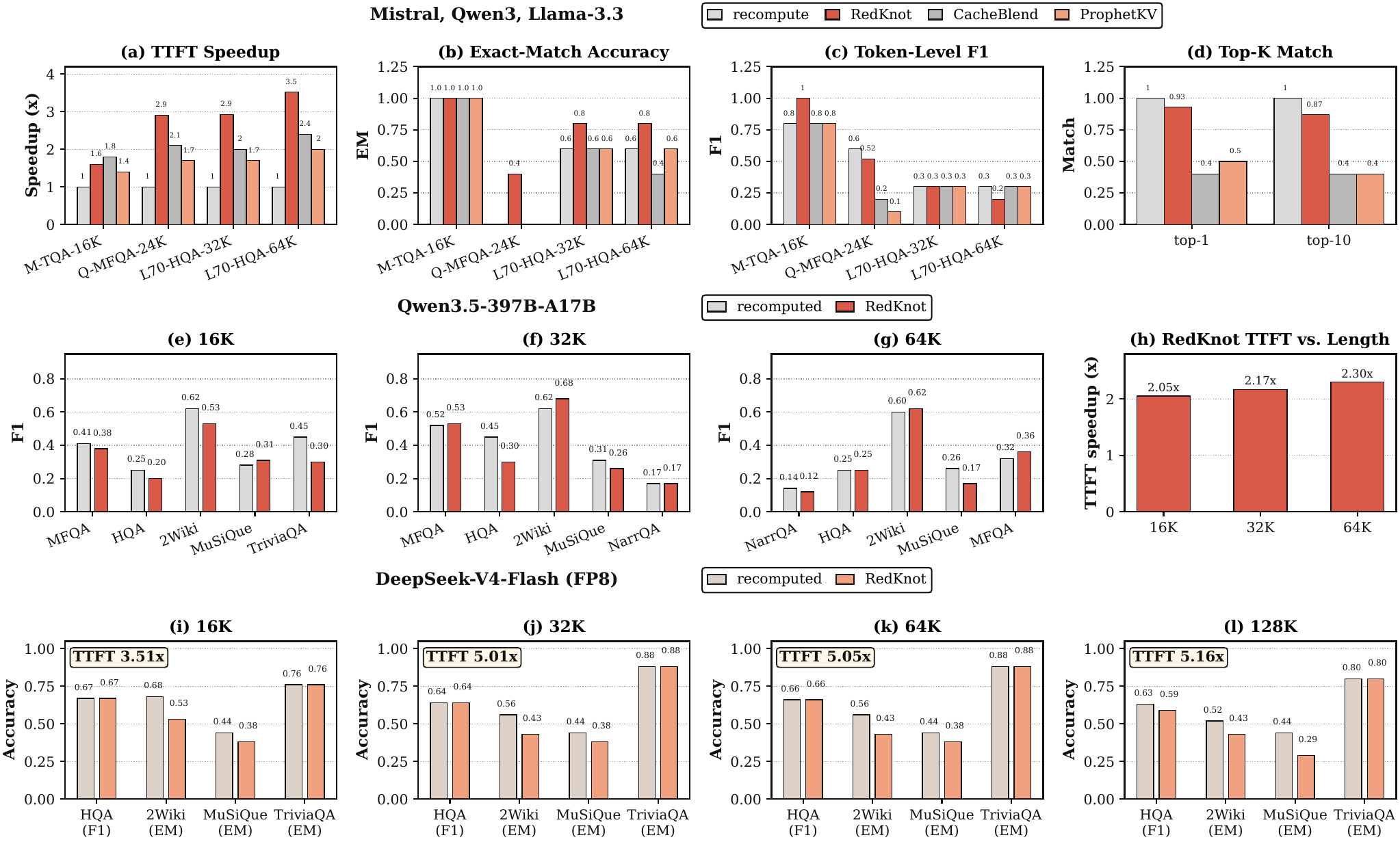} 
  \caption{End-to-end accuracy and TTFT comparison across three model
  families. ~\sysname~achives the per-panel TTFT
  speedup ranging from $3.51\times$ to $5.16\times$. Overall, \sysname\
  preserves accuracy close to full recompute (typically $\ge 95\%$ of the
  dense F1) while delivering $1.4$--$5.2\times$ TTFT speedup, and
  consistently dominates the token-level PIC baselines on the
  quality--latency trade-off.}
  \vspace{-4ex}
  \label{fig:redknot-main-results}
\end{figure*}
\textbf{Hardware and software.}~All experiments run on a single server with
8$\times$~NVIDIA~H800 GPUs (80\,GB HBM each, ${\sim}3.3$\,TB/s HBM bandwidth,
PCIe~Gen5~$\times$16), two Intel Xeon~Platinum~8468V CPUs (96~cores /
192~threads total), 2\,TB DDR memory, and a 14\,TB NVMe RAID-0 local array.
Model weights and LongBench datasets are served from a JuiceFS network mount.
PD-disaggregation experiments use four RDMA RoCE~v2 interfaces with measured
unidirectional bandwidth of ${\sim}200$\,Gbps. The software stack is
Ubuntu~24.04, CUDA~12.9, PyTorch~2.9.1, Triton~3.5.1, HuggingFace
\texttt{transformers}~4.57.1~\cite{wolf2020transformers}, Transformer
Engine~2.10.0, SGLang~\cite{zheng2024sglang}, and a customized
vLLM~0.13.0~\cite{kwon2023vllm} for PD experiments. All methods use the same
container image.\\
\noindent \textbf{Models and offline sparsity configuration.}~We evaluate
Mistral-7B, Qwen3-32B, Llama-3.3-70B, DeepSeek-V4-Flash, and
Qwen3.5-397B-A17B. Before online serving, \sysname\ runs an offline profiling
stage on calibration RAG prompts to decide which heads are \emph{global} or
\emph{local}, the local KV window, and the Sparse-FFN token-selection
thresholds. We report $\rho_g$ as the effective fraction of layer--KV-head
entries that require global or retrieval-style recovery over the evaluated
configuration; local heads keep only a sink/recent window. Unless otherwise
specified, sink size is 4 tokens for GQA models.\\
\indent \emph{Mistral-7B} runs with $\text{TP}=1$, effective
$\rho_g=9.4\%$, and local KV window $W=256$. \emph{Llama-3.3-70B} runs with
$\text{TP}=8$, effective $\rho_g=10.0\%$, and local KV window $W=256$; its
Sparse-FFN configuration keeps the first 20 layers dense, uses
\texttt{mass\_thresh=0.2}, switches to \texttt{mass\_thresh\_deep=0.05}
after layer~60, and always keeps the most recent 512 tokens.
\emph{Qwen3-32B} runs with $\text{TP}=2$ and uses a hybrid attention plan:
the first 48 layers remain full attention because their
\texttt{q\_norm}/\texttt{k\_norm} patterns are high entropy, while the last
16 layers use a sliding window of $W=4096$. Its effective global-head budget
is $\rho_g=9.4\%$ over all layer--head entries; Sparse-FFN keeps the first
5 layers dense, uses \texttt{mass\_thresh=0.2}, switches to
\texttt{mass\_thresh\_deep=0.05} after layer~40, and keeps the most recent
128 tokens.\\
\indent \emph{Qwen3.5-397B-A17B} runs with $\text{TP}=8$ and has 60 layers,
of which 15 are full-attention layers (every fourth layer) and 45 are
GatedDeltaNet-style linear-attention layers. RedKnot sparsifies only the deep
full-attention layers: the first 9 full-attention layers are dense, while the
6 deep full-attention layers use $40\%$ global KV heads and $60\%$ local KV
heads with $W=2048$ and sink size 4, giving an effective global-head budget
of $\rho_g\approx4.3\%$ over the full model. For the linear-attention component,
the first 5 layers are dense; head windows are derived from the 0.95 decay
quantile with safety factor 2.0, minimum window 256, and segment size 2048.
For MoE recovery, layers $\ge24$ use attention-mass sparse routing with
\texttt{mass\_thresh=0.7}; tokens below the threshold skip routed experts and
keep the shared path only. This is the sweet-spot configuration selected on the
397B sweep: it is lossless on the 32K TriviaQA calibration setting while saving
about $52\%$ total compute and giving about $2.07\times$ TTFT speedup.\\
\indent \emph{DeepSeek-V4-Flash} runs with $\text{TP}=8$ in model-quality
experiments and pipeline-parallel serving in the QPS experiments. It uses MLA,
so the physical KV cache has one shared latent KV head even though the attention
module has 64 logical heads. The offline plan marks one logical head per layer as
global/retrieval and the rest as local with a default window of 128 and no sink,
corresponding to an effective retrieval budget of $\rho_g=4.6\%$ under the
indexer top-$k$ selection used in the experiments. Its Sparse-FFN configuration
keeps the first 4 layers dense, uses the native indexer signal
(\texttt{c4\_topk\_lengths\_raw}) as token importance, applies
\texttt{mass\_thresh=0.6}, and always keeps the most recent 256 tokens.\\
\noindent \textbf{Datasets.}~We draw RAG prompts from HotpotQA~\cite{yang2018hotpotqa},
MuSiQue~\cite{trivedi2022musique}, 2WikiMQA~\cite{ho2020constructing},
TriviaQA~\cite{joshi2017triviaqa}, MultiFieldQA~\cite{bai2024longbench},
Qasper~\cite{dasigi2021qasper}, and additional LongBench-style workloads used
in later analyses such as NarrativeQA, GovReport, WikiText, and LCC. Each prompt
concatenates a question with $N_{\text{seg}}$ retrieved passages, including the
gold passage and distractors, with the gold passage randomly placed. We vary the
number of passages and per-passage token budgets to cover contexts from about
8K to 128K tokens, depending on the model and experiment.\\
\noindent \textbf{Baselines and metrics.}~We compare against dense HuggingFace
generation with \texttt{sdpa}, dense FlashAttention-3~\cite{shah2024flashattention3},
SGLang~\cite{sglang2025hicache}, vLLM-style PD serving, and token-level PIC
baselines CacheBlend and ProphetKV (with the recovery ratios stated in the
corresponding figures). Quality is measured by F1, EM, first-token logit cosine,
top-1 agreement, and top-10 overlap against dense recompute. Efficiency is
measured by TTFT, TTFT speedup, analytical prefill FLOPs, KV-transfer bytes,
QPS/GPU, concurrent sessions per GPU, and TTFT coefficient of variation.
For \sysname\ paths, offline segment prefill is excluded from online TTFT and is
reported separately when relevant.
\subsection{Quality and TTFT Comparison}
\label{sec:eval_quality_ttft}
We evaluate the accuracy--latency trade-off of \sysname\ across three
model families that differ in scale and architecture:
the dense small/medium models Mistral-7B, Qwen3-32B, and Llama-3.3-70B;
the large MoE model Qwen3.5-397B-A17B; and the FP8 MoE model
DeepSeek-V4-Flash. For the first group we compare \sysname\ against
dense full recompute and two representative \emph{token-level} PIC
baselines, \emph{CacheBlend} and \emph{ProphetKV}, on four RAG-style
QA workloads (M-TQA-16K, Q-MFQA-24K, L70-HQA-32K, and L70-HQA-64K),
where the panel prefix encodes \emph{model}-\emph{dataset}-\emph{context
length}. For the two large models, where the token-level PIC baselines
do not run reliably at scale, we compare \sysname\ directly against full
recompute across context lengths from 16K to 128K and several LongBench
datasets (HotpotQA, 2WikiMQA, MuSiQue, TriviaQA, MultiFieldQA, NarrativeQA),
with the number of concatenated RAG chunks varying from 4 to 6.
We report accuracy along four dimensions---exact match (EM),
token-level F1, and first-token top-$1$/top-$10$ agreement with the
dense path---together with TTFT speedup over dense prefill. All numbers
are summarized in Figure~\ref{fig:redknot-main-results}.~Overall, \sysname\ achieves a more favorable quality--latency trade-off
than token-level PIC baselines: it preserves accuracy close to full
recompute (typically $\ge 95\%$ of the dense F1) while delivering
$1.4$--$5.2\times$ TTFT speedup, and this advantage grows with context
length.\\
\noindent \textbf{Quality comparison.}
Figure~\ref{fig:redknot-main-results}(b) and (c) show that \sysname\ is
consistently competitive in token-level F1 and clearly stronger in
exact-match accuracy.
On the Llama-3.3-70B HotpotQA cases, \sysname\ improves EM from $0.60$
(dense) to $0.80$ at both 32K and 64K, while CacheBlend and ProphetKV
stay at or below the dense baseline ($0.4$--$0.6$).
On the Qwen3-32B MultiFieldQA case, \sysname\ keeps F1 at $0.52$, close
to the dense $0.6$, whereas ProphetKV collapses to nearly $0.1$.
Even where ProphetKV or CacheBlend remain partially competitive in F1,
their EM is generally lower, indicating that token-level sparse
recomputation often fails to recover the exact answer even when surface
token overlap is nontrivial.
The top-$K$ statistics in Figure~\ref{fig:redknot-main-results}(d)
support the same conclusion: \sysname\ reaches $0.93$ top-$1$ and $0.87$
top-$10$ agreement with the dense path, far above CacheBlend and
ProphetKV ($\le 0.5$), confirming that \sysname\ stays much closer to the
dense next-token distribution.~\sysname\ also generalizes to larger and architecturally different
models. On Qwen3.5-397B-A17B (Figure~\ref{fig:redknot-main-results}(e)--(g)),
\sysname\ tracks full recompute closely across datasets and even slightly
exceeds it on several cases (e.g., F1 of $0.68$ vs.\ $0.62$ on 2WikiMQA
at 32K, and $0.62$ vs.\ $0.60$ at 64K), keeping accuracy at roughly
$95\%$ of the recomputed F1 on average. On DeepSeek-V4-Flash with FP8
(Figure~\ref{fig:redknot-main-results}(i)--(l)), \sysname\ matches the
recomputed baseline almost exactly on HotpotQA and TriviaQA (e.g., F1
$0.67$ and EM $0.76$ at 16K, identical to recompute) and stays within a
small margin on the harder multi-hop datasets up to 128K.~The main reason is that \sysname\ performs \emph{head-aware recovery}
instead of token-level recovery.
Existing PIC baselines such as CacheBlend and ProphetKV decide which
\emph{tokens} should be recomputed or corrected.
However, under multi-head attention, different heads attend to different
token subsets and exhibit different sensitivity to prefix changes.
As a result, token-level methods either recompute too few tokens, which
leaves some head-specific errors uncorrected, or recompute too many
tokens, which weakens the latency benefit.
In contrast, \sysname\ separates global heads from local heads:
prefix-sensitive global heads are explicitly recomputed, while
prefix-robust local heads are largely reused with lightweight repair.
This preserves head-specific sparse structure instead of collapsing all
heads into a single token-level decision, allowing \sysname\ to recover
the dominant attention behavior more faithfully and directly improving
EM, F1, and next-token agreement.\\
\indent \textbf{TTFT comparison.}
Figure~\ref{fig:redknot-main-results}(a), (h), and (i)--(l) show that
\sysname\ also achieves the best TTFT speedup in most long-context
settings, and its advantage becomes more pronounced as the context
grows. On the dense models (Figure~\ref{fig:redknot-main-results}(a)),
\sysname\ reaches $1.6\times$ at M-TQA-16K and rises to $3.5\times$ at
L70-HQA-64K, the largest speedup among all compared methods at that
length, whereas CacheBlend and ProphetKV plateau around $2.0$--$2.4\times$.
The same monotonic trend holds on Qwen3.5-397B-A17B
(Figure~\ref{fig:redknot-main-results}(h)), where the \sysname\ speedup
grows from $2.05\times$ at 16K to $2.17\times$ at 32K and $2.30\times$ at
64K, and is even stronger on the FP8 DeepSeek-V4-Flash
(Figure~\ref{fig:redknot-main-results}(i)--(l)), reaching $3.51\times$ at
16K and up to $5.16\times$ at 128K.~This speed advantage comes from two complementary sources.
First, \sysname\ reduces attention recovery cost through head-aware
execution: only the small set of prefix-sensitive global heads are
recomputed with full-range attention, while most local heads directly
reuse cached KV states or access only bounded visible regions. This
avoids the token-level union effect that forces CacheBlend and ProphetKV
to reprocess a larger set of tokens than any single head actually
requires.
Second, \sysname\ further accelerates prefill through \emph{sparse FFN
recovery}. Token-level PIC baselines mainly optimize the attention path
but keep FFN computation dense; as context grows, this dense FFN
increasingly dominates TTFT and caps their speedup. \sysname\ breaks this
ceiling by applying sparse FFN computation only to important token states
after attention recovery. Reducing both attention-side and FFN-side cost
is why \sysname's TTFT scaling improves with length while the baselines
saturate.\\
\indent \textbf{Why \sysname\ achieves a better quality--latency trade-off.}~The key difference is that \sysname\ aligns its recovery granularity with
the intrinsic structure of LLM inference. At the attention level it
exploits head-level heterogeneity, treating global heads differently from
local heads; at the FFN level it exploits token-selective recovery,
avoiding dense FFN replay for less important token states. By contrast,
token-level PIC baselines operate at a coarser granularity and only
partially exploit attention sparsity, while leaving FFN computation
largely untouched. As a result, they face a less favorable trade-off:
aggressively reducing recomputation hurts accuracy, while preserving
accuracy requires more token recomputation and limits TTFT improvement.
\begin{figure*}
  \centering
  \includegraphics[width=0.98\textwidth]{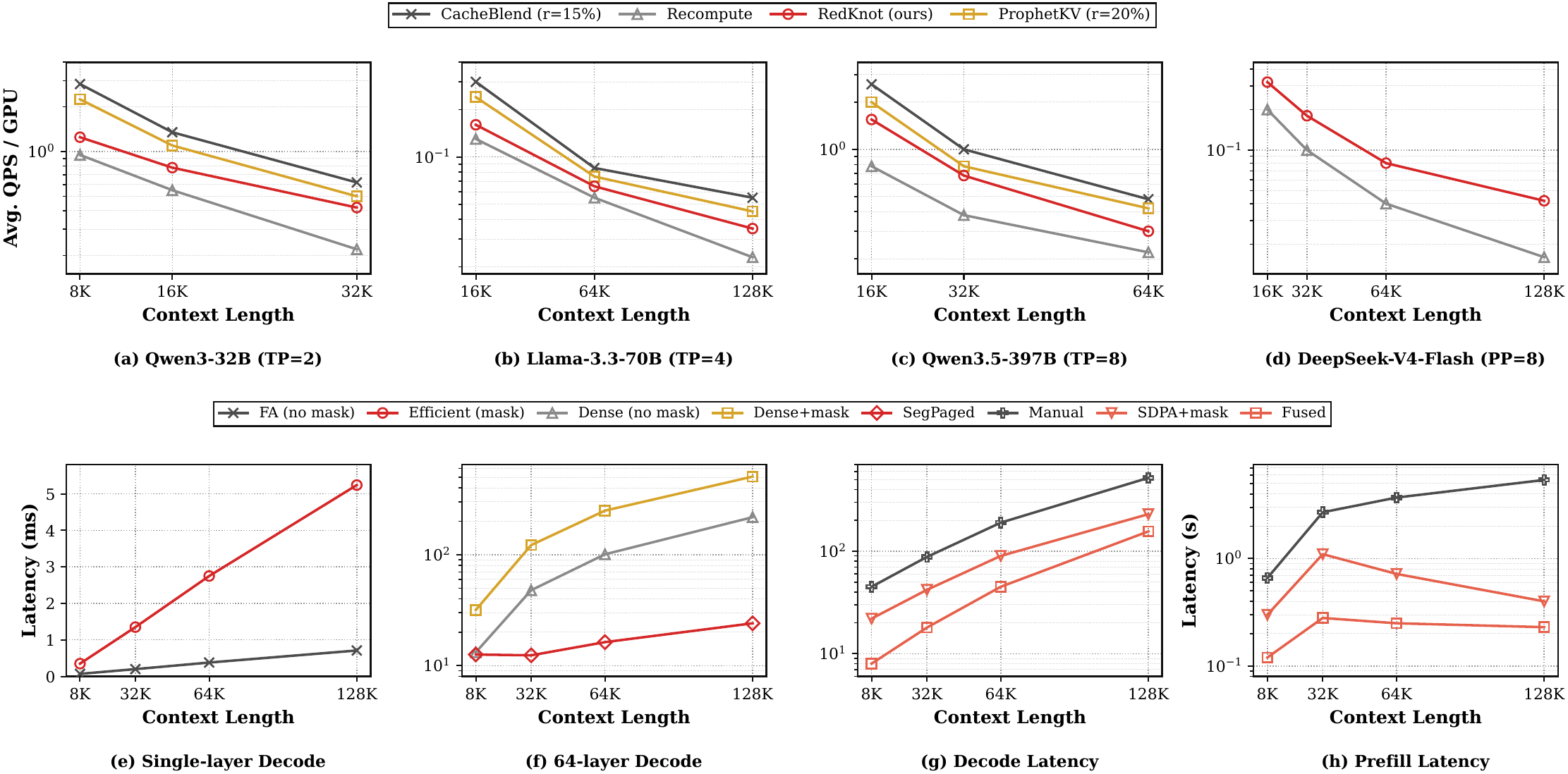} 
  \caption{Throughput and attention-kernel efficiency of \sysname.
  \emph{Top row}~(a)--(d): serving throughput (QPS/GPU, log scale) vs.\
  context length on (a)~Qwen3-32B (TP=2), (b)~Llama-3.3-70B (TP=4),
  (c)~Qwen3.5-397B (TP=8), and (d)~DeepSeek-V4-Flash (PP=8), comparing
  \sysname\ with dense recompute, CacheBlend ($r=15\%$), and ProphetKV
  ($r=20\%$). \emph{Bottom row}~(e)--(h): kernel-isolated latency of
  \SegPagedAttention\ vs.\ masked/dense back-ends for (e)~single-layer
  decode, (f)~64-layer decode, (g)~decode latency (log scale), and
  (h)~prefill latency (log scale); all paths are numerically equivalent
  ($\cos > 0.99998$).}
  \vspace{-3ex}
  \label{fig:redknot-throughput}
\end{figure*}
\begin{figure*}[t]
  \centering
  \includegraphics[width=0.98\textwidth]{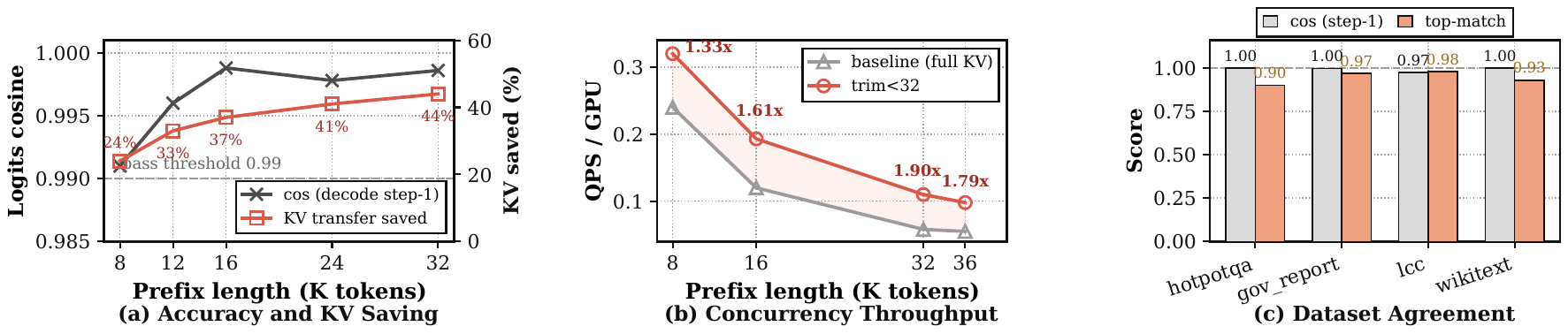} 
  \caption{Prefix multi-head KV compression on Qwen3-32B under PD
  disaggregation. (a)~first-decode-step logit cosine vs.\ the full-KV
  baseline (left axis) and KV-transfer saving (right axis) vs.\ prefix
  length; the dashed line marks the $0.99$ pass threshold.
  (b)~aggregate decode throughput (QPS/GPU) under a fixed KV-memory
  budget, full-KV baseline vs.\ \texttt{trim<32}, with the per-point
  speedup annotated. (c)~per-dataset logit cosine and top-match
  (per-token agreement) against the full-KV output.}
  \label{fig:redknot-prefix-compression}
  \vspace{-3ex}
\end{figure*}
\subsection{Throughput and SegPagedAttention}
\label{sec:eval:throughput}
We now evaluate \sysname\ from two complementary system angles:
end-to-end serving throughput (\cref{fig:redknot-throughput}(a)--(d)),
and the attention-kernel efficiency that ultimately sets the throughput
ceiling (\cref{fig:redknot-throughput}(e)--(h)). The first shows that
head-class KV reuse already improves throughput over dense recompute,
while the second shows that \SegPagedAttention\ removes the remaining
mask overhead that currently limits how much of that benefit is realized.\\
\noindent \textbf{QPS throughput.}~We submit bursts of RAG-style requests and report the average completed
queries per second normalized per GPU (\emph{QPS/GPU}), so that models
served at different parallelism degrees are directly comparable. We sweep
four models at their production parallel configurations---Qwen3-32B
(TP=2), Llama-3.3-70B (TP=4), Qwen3.5-397B (TP=8), and DeepSeek-V4-Flash
(PP=8)---across context lengths from 8K to 128K, comparing dense full
\emph{recompute}, \emph{CacheBlend} ($r=15\%$), \emph{ProphetKV}
($r=20\%$), and \sysname. As shown in
\cref{fig:redknot-throughput}(a)--(d), throughput decreases with context
length for every method, but all KV-reuse methods sustain substantially
higher QPS/GPU than dense recompute, and the advantage grows with length.
On Qwen3.5-397B at 64K, \sysname\ reaches roughly $0.2$~QPS/GPU versus
about $0.05$ for recompute (a $\sim$$4\times$ gain), and a similar
separation holds on Llama-3.3-70B at 128K. On DeepSeek-V4-Flash, where
the token-level PIC baselines do not run reliably at scale, \sysname\
stays well above recompute across all lengths, sustaining about
$0.04$~QPS/GPU at 128K versus $\sim$$0.015$ for recompute.
Compared with the token-level baselines, \sysname\ is competitive but not
always the highest at short contexts: at 8K--16K, CacheBlend and ProphetKV
sit slightly above \sysname\ on the dense models, because their fixed low
recovery ratios recompute only a small token subset, whereas the current
\sysname\ serving backend still expresses head-class sparsity through a
dense KV layout plus an attention mask. As context length grows the curves
converge, since the dense-FFN and mask-fallback costs begin to dominate
every method's prefill. The advantage over recompute is structural---
\sysname\ recomputes only prefix-sensitive global heads while reusing
local-head KV, so the dense baseline alone pays full quadratic attention
on all heads---but the residual gap to token-level PIC at short contexts is
an implementation artifact of the dense+mask backend, which the next set of
measurements isolates and removes.\\
\noindent \textbf{SegPagedAttention: latency reduction.}~We isolate the attention kernel with micro-benchmarks on
Qwen3-32B-shaped layers (64 layers, $H_q=32$, $H_{kv}=8$, $D=128$, bf16,
GQA-4), sweeping 8K/32K/128K context. The head-class layout keeps half of
the KV heads as global heads that read the full context and half as local
heads that retain only a 320-token window (sink plus recent tokens). We
compare a \emph{Dense+mask} layout that encodes head classes with an
additive \texttt{attn\_mask} against \SegPagedAttention, which stores KV
as ragged per-head pages and calls mask-free FlashAttention through a
single fused \texttt{flash\_attn\_varlen\_func}.
The dense mask is the wrong physical interface: PyTorch SDPA can dispatch
to the FlashAttention backend only when the mask is null, so a
materialized \texttt{attn\_mask} forces a slower path whose traffic grows
with the dense context length. As a result
(\cref{fig:redknot-throughput}(e),(f)), Dense+mask decode grows from
$31.8$\,ms at 8K to $506.8$\,ms at 128K, while \SegPagedAttention\ stores
each head at its own live length and stays nearly flat---$12.6$\,ms at 8K,
$12.4$\,ms at 32K, and only $24.1$\,ms at 128K---giving $2.5\times$,
$9.8\times$, and $21.0\times$ decode speedups.
Fusing all heads of a layer into one varlen call further removes per-head
launch overhead (\cref{fig:redknot-throughput}(g)), making fused decode
$2.85$--$3.39\times$ faster than SDPA+mask from 8K to 128K. The prefill
gain is larger still (\cref{fig:redknot-throughput}(h)): \SegPagedAttention\
cuts 64-layer prefill from $0.34$\,s to $0.055$\,s at 8K, $1.35$\,s to
$0.12$\,s at 32K, and $5.42$\,s to $0.23$\,s at 128K---a $6.3\times$,
$11.3\times$, and $23.3\times$ speedup---because prefill is dominated by
the quadratic attention work that dense+mask still performs over all
heads. The speedup grows with length precisely because the dense baseline
pays full $L$ on every head, whereas \SegPagedAttention\ pays full $L$
only on the small set of global heads.\\
\noindent \textbf{SegPagedAttention: bandwidth and throughput utilization.}~The latency reduction is fundamentally a memory-bandwidth effect.
A dense PagedAttention-style layout can express which tokens a head should
ignore, but it cannot stop the GPU from loading and scheduling work for
those tokens, so it keeps streaming HBM traffic proportional to the full
$[B,H,L,D]$ tensor on every head. \SegPagedAttention\ changes the
contract: each head owns a compact page list, the kernel consumes ragged
per-head lengths directly, and no additive mask is built, so local heads
move only their 320-token window across the memory hierarchy while global
heads move the full context. This converts the saved KV bytes into saved
bandwidth, which is what keeps decode latency flat and prefill cost
near-linear in the number of \emph{global}-head tokens rather than total
tokens. The effect is visible as token throughput: the SDPA+mask path
falls from $5.9$K tok/s at 8K to $1.5$K tok/s at 32K, retaining only
$\sim$$25\%$ of its 8K throughput, whereas \SegPagedAttention\ falls from
$37.4$K to $17.1$K tok/s, retaining $46\%$ while remaining $6.3\times$
faster at 8K and $11.4\times$ faster at 32K. The residual degradation
comes only from global heads, whose KV still grows with $L$; local heads
contribute constant-length traffic and the fused varlen kernel keeps
execution mask-free.
\subsection{Prefix Compression}
\label{sec:eval:prefix}
Prefix multi-head compression targets the PD-disaggregated setting, where
a prefill node produces the full prefix KV cache and ships it to a decode
node; the two dominant costs are the prefix$\rightarrow$decode KV-transfer
volume and the decode node's KV memory, which bounds concurrency. The idea
follows \sysname's head classes: \emph{global}/retrieval heads keep the
whole prefix KV, while \emph{local} heads keep only a bounded
sink-plus-window region and evict the middle. Eviction is \emph{true
eviction}---the trimmed KV is never stored and never enters the
softmax---rather than zero-filling, which would still contribute
$\exp(0)$ mass. We evaluate on Qwen3-32B (64 layers, 8 KV heads,
$D=128$, GQA, native context $40{,}960$, bf16) on two GPUs. Because
Qwen3-32B is a \emph{dense} model with no native sliding-window mask,
aggressive all-layer trimming collapses; the accuracy-safe operating
point is \texttt{trim<32}, i.e.\ trimming local heads only in the first
$32$ of $64$ layers, with window $W{=}4096$ and sink $=128$. All results
use this single configuration. We report three views
(\cref{fig:redknot-prefix-compression}): (a)~accuracy and KV-transfer
saving vs.\ prefix length, (b)~memory-bound concurrency throughput, and
(c)~cross-dataset accuracy.\\
\indent \emph{Accuracy and KV saving} (\cref{fig:redknot-prefix-compression}(a)):
the first-decode-step logit cosine against the full-KV baseline stays
above the $0.99$ pass threshold at every prefix length
($0.9911$ at 8K, $0.9988$ at 16K, $0.9987$ at 32K), while the
KV-transfer saving rises monotonically from $24\%$ at 8K to $44\%$ at
32K. \emph{Concurrency throughput}
(\cref{fig:redknot-prefix-compression}(b)): under a fixed
$\sim$$46$\,GiB KV budget, compression increases the maximum concurrent
batch (e.g.\ $5\rightarrow10$ at 32K) and lifts aggregate decode QPS/GPU
from $0.057$ to $0.108$ at 32K, a $1.90\times$ speedup, with the gain
growing from $1.33\times$ at 8K to $1.90\times$ at 32K. \emph{Cross-dataset
accuracy} (\cref{fig:redknot-prefix-compression}(c)): logit cosine stays
$\ge 0.974$ across HotpotQA, GovReport, LCC, and WikiText, and per-token
top-match remains high ($0.90$--$0.98$), confirming faithful reproduction
of the full-KV output across QA, summarization, code, and language
modeling.\\
\indent \textbf{Analysis.}~Two structural reasons explain the results.
First, accuracy is preserved because local heads, by definition, attend
almost entirely to the sink and recent window at decode time, so the
evicted middle of the prefix carries little of their attention mass;
keeping the later $32$ ``information-extraction'' layers at full KV
protects the heads that do need long-range context, which is why the
cosine never drops below the pass threshold. Second, the KV-transfer
saving grows with prefix length because the fixed sink-plus-window region
occupies a shrinking fraction of a longer prefix, so longer contexts---
exactly where PD disaggregation matters most---benefit the most.
The dominant win, however, is concurrency rather than single-stream
latency: decode is memory-bound, so a smaller per-request KV lets more
requests share one decode GPU, and the throughput gain (up to $1.90\times$)
slightly exceeds the batch-size gain because shorter per-request KV also
speeds up attention. Single-stream latency changes little ($\sim$$1.1\times$),
confirming that the benefit is fundamentally driven by concurrency under a
fixed memory budget. The main caveat is task awareness: on
retrieval-heavy prompts the answer-bearing token may lie in the evicted
middle, so deployment should keep a retrieval-head allow-list or apply
the compression with task awareness.
\subsection{KV Cache Lifecycle Management}
\label{sec:eval:lifecycle}
\begin{figure}[t]
  \centering
  \includegraphics[width=0.98\linewidth]{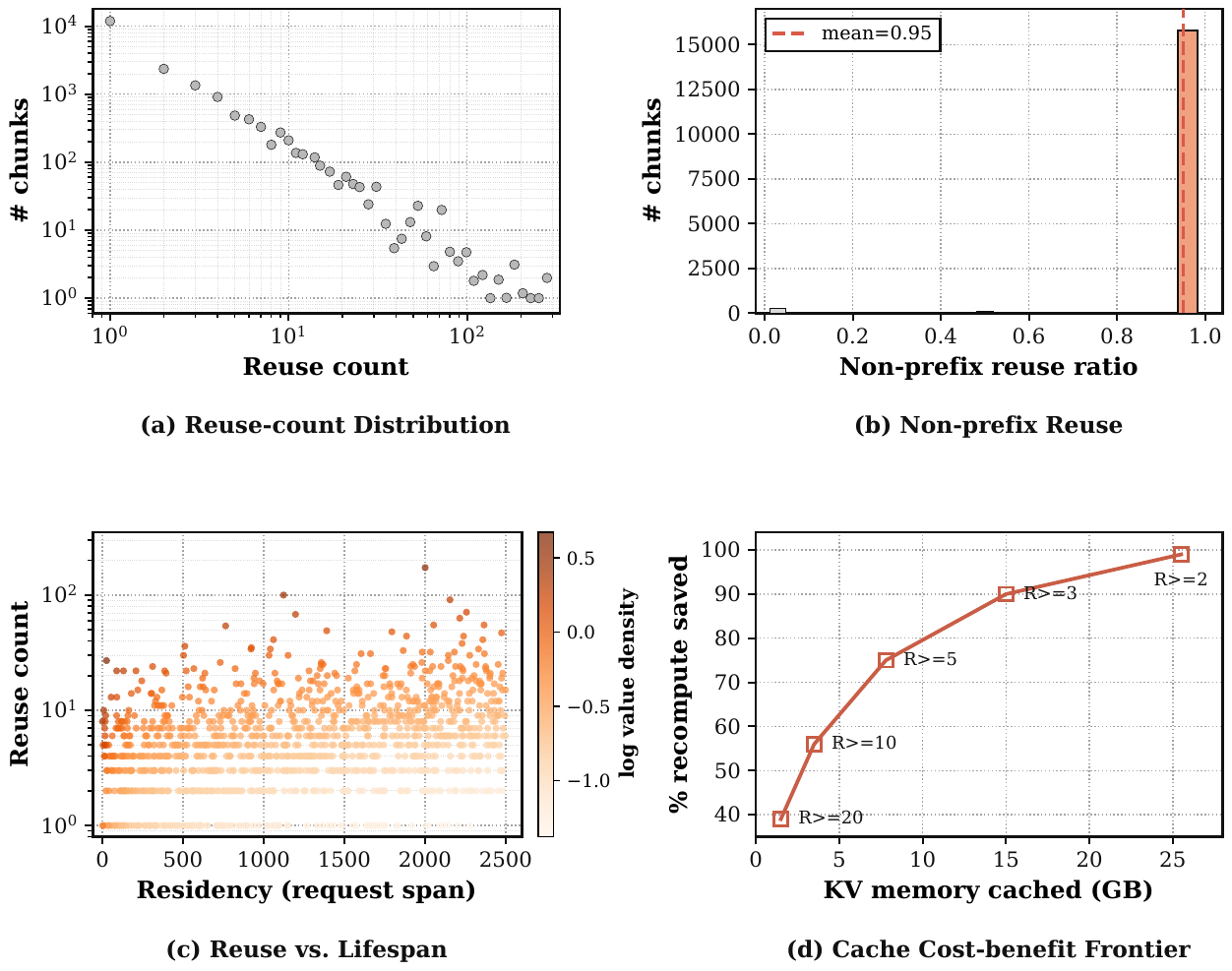} 
  \caption{Chunk-level KV reuse on the MuSiQue stream
  ($2417$ questions, $48{,}315$ chunk accesses, $17{,}629$ unique
  passages). (a)~reuse-count distribution per chunk (log--log).
  (b)~fraction of each chunk's reuse that comes from non-prefix
  positions, with mean $0.95$. (c)~reuse count vs.\ residency (the
  request span over which a chunk stays live), colored by
  log value density. (d)~recompute saved vs.\ the KV memory needed to
  cache all chunks whose reuse count is at least $R$.}
  \vspace{-3ex}
  \label{fig:redknot-kv-lifecycle}
\end{figure}
~Caching every chunk's KV for later reuse is appealing but does not scale:
on MuSiQue, $62\%$ of chunks are never reused, yet a ``cache-all'' policy
would need $524$\,GB of KV to hold them (DeepSeek-V4 MLA). To decide what
to cache and for how long, we first characterize how KV is actually
reused. We replay the real MuSiQue request stream---$2417$ questions,
$48{,}315$ chunk accesses over $17{,}629$ unique passages---and, for each
chunk, record its reuse count, the share of reuse that is \emph{non-prefix}
(i.e.\ the chunk is reused at a position other than a shared prefix), and
its residency, defined as the request span between its first and last
access. \cref{fig:redknot-kv-lifecycle} summarizes these statistics and,
in panel~(d), the resulting trade-off between cached KV memory and saved
recomputation.\\
\indent 
~The reuse-count distribution in
\cref{fig:redknot-kv-lifecycle}(a) is heavy-tailed: most chunks are seen
only once or twice, while a small set is reused tens to hundreds of times.
\cref{fig:redknot-kv-lifecycle}(b) shows that reuse is almost entirely
non-prefix---the per-chunk non-prefix ratio concentrates near $0.95$ on
average---so prefix caching alone captures little of the available reuse.
\cref{fig:redknot-kv-lifecycle}(c) plots reuse against residency: highly
reused chunks tend to stay live across long request spans, whereas the
many single-use chunks appear briefly and never return, so reuse and
lifespan are correlated rather than independent.
\cref{fig:redknot-kv-lifecycle}(d) turns this into a cost--benefit
frontier. Caching only chunks reused at least $R$ times trades memory for
saved recompute: admitting chunks with $R\!\ge\!20$ already saves $39\%$
of recomputation using $1.5$\,GB, $R\!\ge\!5$ saves $75\%$ at $7.8$\,GB,
and $R\!\ge\!2$ saves $99\%$ but needs $25.5$\,GB. The curve bends
sharply, so most of the benefit is reached well before the memory cost
becomes large.\\
\indent \textbf{Analysis.}~These statistics explain when KV cache should be produced and how it
should be managed. Because the majority of chunks are one-shot
(\cref{fig:redknot-kv-lifecycle}(a)), materializing and storing KV on
first sight is wasteful; KV should instead be produced only after a chunk
has proven worth caching, which motivates an admission gate that promotes
a chunk only once its reuse count crosses a threshold. The diminishing
returns in \cref{fig:redknot-kv-lifecycle}(d) make this concrete: moving
from $R\!\ge\!2$ to $R\!\ge\!5$ gives up only $24$ points of saved
recompute but cuts the memory footprint by more than $3\times$, so a
modest threshold removes most of the storage cost while keeping most of
the benefit. Once a chunk is admitted, the correlation between reuse and
residency in \cref{fig:redknot-kv-lifecycle}(c) guides eviction and
expiry: chunks that are both frequently reused and long-lived are the
ones worth keeping, while chunks that have gone idle can be released
without losing future hits. The non-prefix dominance in
\cref{fig:redknot-kv-lifecycle}(b) further shows that this management must
operate at the granularity of arbitrary chunks rather than shared
prefixes, since prefix-based reuse would miss most of the traffic.
Together, these observations motivate a lifecycle policy that admits
chunks by frequency, evicts by a combination of reuse and recency, and
expires idle entries---keeping the hot long tail resident while avoiding
the storage blowup of caching everything.
\subsection{Other System-Level Benefits}
\label{sec:eval:system}
\begin{figure*}
  \centering
  \includegraphics[width=0.98\textwidth]{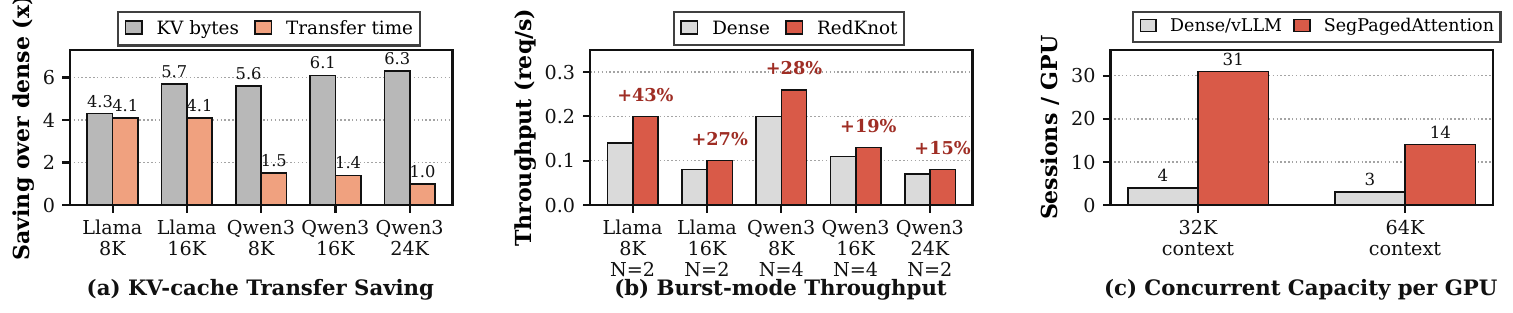} 
  \caption{System-level effects of head-class KV sparsity.
  (a)~KV-cache transfer saving over dense PD disaggregation, separated
  into transferred bytes and wall-clock transfer time, across Llama-3.3-70B
  and Qwen3-32B at 8K--24K. (b)~burst-mode throughput (req/s) of dense
  vs.\ \sysname\ for bursts of $N$ concurrent requests, annotated with the
  relative gain. (c)~concurrent sessions per GPU under dense vLLM-style KV
  storage vs.\ \sysname\ with SegPagedAttention at 32K and 64K context.}
  \label{fig:redknot-other-results}
\end{figure*}
Beyond answer quality, TTFT, and prefill compute, long-context serving is
limited by three further bottlenecks: the volume of KV that must cross the
prefill--decode (PD) boundary, the request throughput under bursty load,
and the number of sessions that fit in GPU memory. We measure all three on
the 8$\times$H800 testbed. The PD experiments use Qwen3-32B and
Llama-3.3-70B on vLLM~0.13.0 with the prefill and decode pools placed on
separate GPU groups; the produced KV cache is transferred from the prefill
side to the decode side before the first token is generated. To match
PagedAttention as the baseline, these runs do not yet enable
SegPagedAttention. For throughput we submit bursts of $N$ simultaneous RAG
requests and report completed requests per second, and for capacity we
count how many concurrent sessions fit in GPU memory under dense
vLLM-style KV storage versus \sysname\ with SegPagedAttention, where the
per-head KV sparsity is physically materialized.

Because head-class sparsity removes KV that local heads never read, it
directly shrinks the payload that must cross the PD boundary
(\cref{fig:redknot-other-results}(a)). On Llama-3.3-70B the transferred KV
volume drops by $4.3\times$ at 8K and $5.7\times$ at 16K, and because the
tensors are large, this byte reduction converts almost one-to-one into
time, cutting transfer latency by $4.1\times$ in both cases. On Qwen3-32B
the byte saving is even larger, $5.6\times$ to $6.3\times$ from 8K to 24K,
but the transfer-time speedup is only $1.0\times$ to $1.5\times$: the
per-request tensors are smaller, so once the bytes are compressed the
remaining time is dominated by launch, packing, and link-setup overheads
rather than by the data itself. The byte saving stays stable across both
models because it is structural. \sysname\ partitions KV by head---global
heads keep the full context, local heads keep only their sink and recent
window, and retrieval heads keep only selected tokens---whereas dense PD
serving ships the entire $[B,H,L,D]$ tensor even for heads that will never
read most positions.~The same KV reduction also improves end-to-end throughput under bursty
load, though more modestly (\cref{fig:redknot-other-results}(b)). On
Llama-3.3-70B throughput rises from $0.14$ to $0.20$~req/s at 8K
($+43\%$) and from $0.08$ to $0.10$~req/s at 16K ($+27\%$); on Qwen3-32B
it rises from $0.20$ to $0.26$~req/s at 8K ($+28\%$), from $0.11$ to
$0.13$~req/s at 16K ($+19\%$), and by $15\%$ at 24K. These gains are
smaller than the isolated TTFT and FLOP reductions because burst
throughput also includes request scheduling, KV packing, transfer,
decode, and runtime orchestration, all of which stay in the critical path.
The gain also shrinks as context grows, which exposes a limitation of this
backend: it still expresses per-head sparsity through a dense KV layout
plus an attention mask, so it saves transfer bytes but keeps paying the
SDPA mask penalty discussed in \cref{sec:eval:throughput}. As the context
lengthens, that masked-attention cost absorbs a larger share of the
benefit, leaving the algorithm's KV reduction only partly realized.\\
\indent Materializing the same sparsity physically with SegPagedAttention removes
that limitation and turns the KV saving into capacity
(\cref{fig:redknot-other-results}(c)). Dense vLLM-style serving is
typically memory-bound rather than compute-bound: once per-session KV
fills the GPU, no further requests can be admitted even when compute is
idle. With per-head KV storage, the number of concurrent sessions per GPU
grows from $4$ to $31$ at 32K context ($7.8\times$) and from $3$ to $14$
at 64K ($4.7\times$), which under a conservative pipelined serving model
projects to $3.4\times$ to $3.9\times$ higher capacity-bound throughput.
A dense layout can mark which tokens a head should ignore, but it still
reserves memory for every head at every position; SegPagedAttention
instead allocates pages per head, so local heads occupy only their short
windows while global heads hold full-context pages. For long-context RAG
this changes the binding constraint from how many full dense KV caches fit
in HBM to how many compact per-head caches fit, which is why the capacity
gain exceeds the single-request throughput gain.
\subsection{Sparse Denoising for Long-Context Attention}
\label{sec:eval:sparse-denoise}
\begin{figure*}[t]
  \centering
  \includegraphics[width=0.98\textwidth]{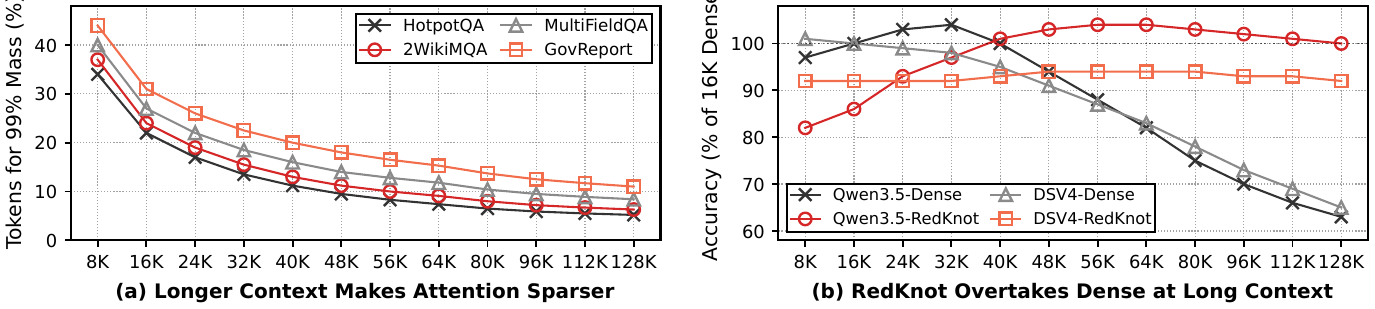}
  \caption{Sparse denoising becomes more useful as context grows.
  (a)~On DeepSeek-V4-Flash, the fraction of tokens needed to cover
  $99\%$ attention mass drops with context length across HotpotQA,
  2WikiMQA, MultiFieldQA, and GovReport. (b)~On Qwen3.5-397B and
  DeepSeek-V4-Flash, dense accuracy degrades under long-context noise,
  while \sysname\ stays stable and overtakes dense at longer contexts.}
  \label{fig:redknot-sparse-denoise}
\end{figure*}

We next isolate a different aspect of long-context behavior: not all
tokens in a long prompt are equally useful for the next-token decision.
As context length grows, retrieved passages, repeated boilerplate, and
task-irrelevant spans introduce increasing amounts of attention noise.
This experiment tests whether the sparse execution used by \sysname\ can
act as a denoising mechanism rather than only as a compute-saving
mechanism. We evaluate Qwen3.5-397B-A17B and DeepSeek-V4-Flash on the
same long-context QA and document workloads used earlier, including
HotpotQA, 2WikiMQA, MultiFieldQA, and GovReport. Context lengths range
from 8K to 128K tokens. For the sparsity measurement, we compute the
minimum fraction of tokens required to cover $99\%$ of the attention mass
in sparse-eligible layers. For the accuracy measurement, we compare dense
full recomputation with \sysname\ at the same context length and report
accuracy normalized to each model's dense 16K result, so that the two
model families can be shown on the same axis.~Figure~\ref{fig:redknot-sparse-denoise}(a) shows that attention becomes
increasingly concentrated as the prompt grows. At short context, different
datasets require a substantial fraction of tokens to preserve $99\%$ of
the mass, but this fraction falls quickly with length. The trend is
consistent across tasks, while the absolute level depends on the task
type. Retrieval-heavy QA tasks such as HotpotQA and 2WikiMQA become the
sparsest because the answer evidence is concentrated in a small subset of
retrieved passages. MultiFieldQA is less sparse, and GovReport remains the
densest because summarization-style inputs distribute useful evidence more
widely across the document. This separation is important: \sysname\ does
not assume a fixed global sparsity ratio, but benefits from the fact that
all tasks become more selective as context length increases.~Figure~\ref{fig:redknot-sparse-denoise}(b) links this sparsity trend to
accuracy. Dense recomputation initially has an advantage at short context
because it preserves every token interaction, including weak long-range
signals. As the context becomes longer, however, dense attention also
propagates more distractor tokens and irrelevant passages through the
model. Its normalized accuracy therefore drops after the medium-length
regime. \sysname\ follows the opposite pattern: it is slightly lower at
short context, where there is little noise to remove, but remains stable
as the prompt grows because sparse recovery suppresses low-value token
states while retaining the high-mass attention structure. On both
Qwen3.5-397B and DeepSeek-V4-Flash, the RedKnot curves cross the dense
curves in the long-context region, showing that sparsity can improve
quality rather than merely preserve it.\\
\indent The result explains why long-context acceleration and accuracy need not be
in conflict. When the prompt is short, dense computation is a reasonable
default because most tokens are still potentially useful. Once the context
expands, the marginal tokens increasingly behave like noise: they consume
attention bandwidth and FFN compute, and they can also perturb the model's
answer distribution. By selecting the heads and token states that carry
most of the useful mass, \sysname\ removes this long-context noise from
the recovery path. The same sparsity that reduces computation therefore
also improves the signal-to-noise ratio of the recovered representation,
which is why the benefit becomes larger at 64K--128K contexts than at
8K--16K contexts.

\section{Future Work}\label{sec:future}
The results of \sysname\ suggest that the next generation of inference
engines should not be organized around dense layers, dense sequences, or
prefix-only cache hits. Those abstractions were a good fit for short
prompts and dense prefill, but they hide the structure that dominates
long-context RAG and agent workloads: different heads need different
context ranges, different chunks have different reuse lifetimes, and
different token states contribute very different amounts of useful signal.
The main future direction is therefore not a single new kernel or cache
optimization, but a new serving contract in which sparsity, reuse, and
validity are visible to the whole engine.\\
\indent \textbf{\emph{Head-Aware KV as a First-Class Engine Object.}}
\label{sec:future:head-aware-kv}
Current serving engines usually expose KV as a rectangular layer-level
tensor. This makes scheduling and memory management simple, but it forces
all heads in a layer to share the same physical context length. Our
experiments show that this is the wrong unit for long-context serving:
only a small set of global or retrieval heads need full-context access,
while most local heads consume only a sink/recent window. \sysname\ uses
this observation algorithmically through head-aware recovery, and
\SegPagedAttention\ turns it into a physical layout by giving each head a
compact page list. The natural next step is to make such per-$(layer,
head)$ KV objects native to the engine rather than implemented as a
special path.~In such an engine, the cache manager would allocate pages according to
each head's live range, the attention API would accept ragged per-head
lengths by default, and the scheduler would reason about heterogeneous
head costs. Global heads are bandwidth-heavy and should remain close to
the GPU; local heads are cheaper and can be compressed, tiered, or
evicted more aggressively. Under GQA, the KV head and its query-head
fanout should also become an explicit scheduling object. This would
remove the current impedance mismatch where an algorithm discovers
per-head sparsity, but the storage layer re-expands it into dense KV, the
kernel consumes it through an attention mask, and the scheduler accounts
for it as if every head were equally expensive. The gains observed from
\SegPagedAttention, prefix compression, and concurrent-session capacity
are all early evidence for the same direction: once sparsity is per-head,
the engine should be per-head as well.\\
\indent \textbf{\emph{Position-Independent KV as the Default Cache Contract.}}
\label{sec:future:pic-contract}
Prefix caching is too narrow for RAG and agent workloads. A document
chunk, tool output, or code block may reappear under a different query,
after a different prefix, or in a different retrieval order. A prefix
cache treats these cases as misses, even though the content has already
been processed. \sysname\ shows that position-independent reuse is
possible, but also that the cache object cannot be a single dense tensor
with a binary hit/miss state. Once a chunk moves behind a new prefix, some
heads remain reusable, some heads require recovery, and some token states
are worth recomputing while others behave like noise.~A future engine should therefore expose a content-addressed, head-aware
cache object. Its metadata would include per-head validity, recovery
cost, compressed footprint, and expected reuse frequency. Cache admission
would prioritize chunks that are likely to recur, not merely long shared
prefixes. Eviction would consider both bytes and recovery value, as in
our KV lifecycle analysis. PD disaggregation would transfer only the
head-class payload needed by the decode side, rather than shipping a dense
layer tensor. This would make position-independent cache reuse a native
cache-system primitive instead of an optimization layered on top of prefix
caching.\\
\indent \textbf{\emph{Noise-Aware Scheduling Across Attention and FFN.}}
\label{sec:future:noise-aware-scheduling}
The sparse-denoising results point to another future engine responsibility:
the runtime should decide not only what can be skipped, but what should be
skipped for quality. In long-context settings, dense computation can
propagate distractor evidence through attention and FFN layers. \sysname\
already uses sparse FFN recovery and indexer-guided token selection to
avoid replaying low-value token states, and the long-context crossover
experiment shows that this can improve accuracy rather than merely
preserve it. A next-generation engine should generalize this idea into a
noise-aware scheduler that treats compute as a budgeted resource assigned
to high-signal heads, tokens, and chunks.~This raises several open questions. The scheduler needs online signals
that are cheap enough to compute during serving but reliable enough to
predict future value. It should adapt thresholds to task type, since
retrieval QA, summarization, code, and tool-use traces have different
information densities. It should also coordinate attention sparsity with
FFN sparsity, because removing attention noise but replaying dense FFN on
all token states leaves a large part of the long-context cost intact.
RedKnot's current implementation provides the first version of these
signals---head classes, indexer mass, sparse-FFN selection, and lifecycle
reuse statistics---but a full engine would make them part of a unified
runtime policy.\\
\indent \textbf{\emph{Toward a Unified Sparse Serving Stack.}}
\label{sec:future:unified-stack}
The common theme is that future inference systems should avoid repeatedly
translating sparse decisions back into dense interfaces. RedKnot currently
demonstrates the pieces separately: head-aware recovery decides which
heads need full context; SegPagedAttention stores and executes the
resulting ragged layout; prefix compression reduces decode-side KV
footprint; lifecycle management decides which reusable chunks deserve KV;
and sparse denoising shows when skipping low-value token states improves
quality. The next step is to integrate these pieces into one serving
stack where cache layout, kernel dispatch, network transfer, admission,
eviction, and scheduling all share the same sparse metadata.~Such an engine would change the performance model of long-context
serving. Instead of asking how fast a dense prefill can be executed, it
would ask which parts of the context are worth materializing, which heads
can consume them, where their KV should live, and whether recomputing them
adds signal or noise. We view this as the central systems challenge for
long-context inference: making sparsity a first-class runtime abstraction
rather than a collection of local optimizations.
\section{Conclusion}
\vspace{-1ex}

\sysname\ revisits position-independent KV cache reuse by aligning its
recovery, compute, and storage granularities with the per-head sparsity
structure of the workload. ~It recovers cached KV at the granularity of
attention heads rather than tokens, materializes per-head sparsity through
\SegPagedAttention\ so that every head stays on the FlashAttention fast path,
and applies token-level Sparse FFN to attack the short-context FFN bottleneck
that no attention-side technique can reach. ~Across three models, six QA
datasets, and context lengths from 8\,K to 128\,K, \sysname\ delivers up to
3.54$\times$ TTFT speedup and 4.7--7.8$\times$ more concurrent sessions per GPU while
cutting prefill FLOPs by 79.5\%, with end-to-end accuracy matching or
exceeding the dense baseline. 
\bibliographystyle{plain}
\bibliography{refs}
\end{document}